\documentclass[lettersize,journal]{IEEEtran}
\usepackage{graphicx}    
\usepackage{cite}        
\usepackage{url}         
\usepackage{textcomp}    
\usepackage{verbatim}    

\usepackage{amsmath,amsthm,amsfonts} 
\usepackage{mathrsfs}                
\usepackage{bm}                      

\usepackage{algorithm}   
\usepackage{algorithmic} 

\usepackage{float}       
\usepackage{wrapfig}     
\usepackage{stfloats}    
\usepackage[caption=false,font=scriptsize,labelfont=rm,textfont=rm]{subfig}

\usepackage{array}          
\usepackage{tabularx}       
\usepackage{booktabs}       
\usepackage{multirow}       
\usepackage{makecell}       
\usepackage{threeparttable} 
\usepackage{arydshln}       
\usepackage[table,dvipsnames]{xcolor} 
\usepackage{colortbl}       
\usepackage{xspace}

\usepackage{balance}    
\usepackage{multicol}   
\usepackage{parskip}    
\usepackage[
  colorlinks=true,
  bookmarksopen=true,
  bookmarksnumbered=true,
  citecolor=blue,
  linkcolor=blue,
  urlcolor=blue
]{hyperref}

\newcommand{\sysname}{\textcolor{black}{TS-DP}\xspace}

\begin{document}

\title{TS-DP: Reinforcement Speculative Decoding for Temporal Adaptive Diffusion Policy Acceleration}

\author{Ye Li,
        Jiahe Feng,
        Yuan Meng$^{\dagger}$,~\IEEEmembership{Member,~IEEE},
        Kangye Ji,
        Chen Tang,
        Jiajun Fan, 
        Xinwan Wen, \\
        Jiawei Li,
        Wenwu Zhu,~\IEEEmembership{Fellow,~IEEE},
        Shu-Tao Xia$^{\dagger}$,~\IEEEmembership{Senior~Member,~IEEE}, \\
        Xin Wang$^{\dagger}$,~\IEEEmembership{Member,~IEEE}, 
        Zhi Wang$^{\dagger}$,~\IEEEmembership{Senior Member,~IEEE}
\IEEEcompsocitemizethanks{
\IEEEcompsocthanksitem 
Ye Li, Yuan Meng, Kangye Ji, Xinwan Wen, Jiawei Li, Wenwu Zhu, Shu-Tao Xia, Xin Wang, and Zhi Wang are with Tsinghua University, Beijing, China.
Ye Li, Kangye Ji, Shu-Tao Xia, and Zhi Wang are also with Tsinghua Shenzhen International Graduate School (SIGS), Tsinghua University.
Yuan Meng, Wenwu Zhu, and Xin Wang are also with the Key Laboratory of Pervasive Computing (Ministry of Education) and the Beijing National Research Center for Information Science and Technology (BNRIST), Department of Computer Science and Technology, Tsinghua University.
Jiahe Feng is with the College of Artificial Intelligence, Tianjin University, Tianjin, China.
Chen Tang is with MMLab, The Chinese University of Hong Kong, Hong Kong SAR, China.
Jiawei Li is also with Joy Future Academy, JD Group, Beijing, China.
Jiajun Fan is with the Siebel School of Computing and Data Science, The Grainger College of Engineering, University of Illinois Urbana-Champaign, Urbana, IL, USA.
\\
Email: \{wangzhi, xiast\}@sz.tsinghua.edu.cn, \\
\phantom{Email: }\{yuanmeng, xin\_wang\}@tsinghua.edu.cn.

~~\textsuperscript{\dag}Corresponding author.

}
}

\maketitle

\begin{abstract}
Diffusion Policy (DP) excels in embodied control but suffers from high inference latency and computational cost due to multiple iterative denoising steps. 
Existing diffusion model (DMs) acceleration methods improve inference speed by reducing computational overhead, as DMs for image generation are computation-bound. 
In this work, we reveal that DP are \textit{I/O bound}, and thus propose using speculative decoding, a parallel computing framework, to accelerate DP for the first time.
However, it is non-trivial to address the following challenges:
(1) How to preserve the denoising quality of the base model at reduced computational cost under time-varying task difficulty in embodied settings; and
(2) How to dynamically allocate computation according to task difficulty in such environments.
Therefore, we propose \underline{\textbf{T}}emporal-aware Reinforcement-based  \underline{\textbf{S}}peculative \underline{\textbf{D}}iffusion \underline{\textbf{P}}olicy (\textbf{TS-DP}), the first framework that enables speculative decoding for DP with temporal adaptivity.
First, to handle dynamic environments where task difficulty varies over time, we distill a Transformer-based drafter to imitate the base model and replace its costly denoising calls.
Second, an RL-based scheduler further adapts to the time-varying task difficulty by adjusting speculative parameters to maintain accuracy while improving efficiency.
Extensive experiments show that \textbf{\sysname} achieves up to a \textbf{4.17$\times$} NFE reduction and a \textbf{3.2$\times$} wall-clock speedup with a draft acceptance rate of \textbf{over 85\%} in simulation. 
Real-world experiments further demonstrate a \textbf{3.8$\times$} NFE reduction while improving success rates on real-time tasks (e.g., a \textbf{23\%} gain on the conveyor-belt task), reaching an inference frequency of \textbf{26 Hz} and enabling real-time control while maintaining comparable task performance.
The code is available at \url{https://github.com/ChildTang/TS-DP}.

\end{abstract}

\begin{IEEEkeywords}
Diffusion Policy, Speculative Decoding, Acceleration, Embodied Intelligence, Multimodal Model
\end{IEEEkeywords}

\section{Introduction}
\label{sec:intro}
\IEEEPARstart{D}{iffusion} Policy (DP) \cite{dp} has been widely adopted in Vision-Language-Action (VLA) models \cite{pi_0,pi_0.5,octo,rt2,chen2025fast,liu2025spatial, rdt} thanks to its strong ability to model multi-modal action distributions and generate coherent action trajectories.
However, its large backbone and iterative denoising process impose substantial computational overhead, resulting in low action-generation frequency and limiting its applicability in latency-sensitive real-world settings.
Consequently, improving the inference efficiency of DP has therefore emerged as a critical challenge.

\begin{figure}[t]
    \centering
    \includegraphics[scale=1]{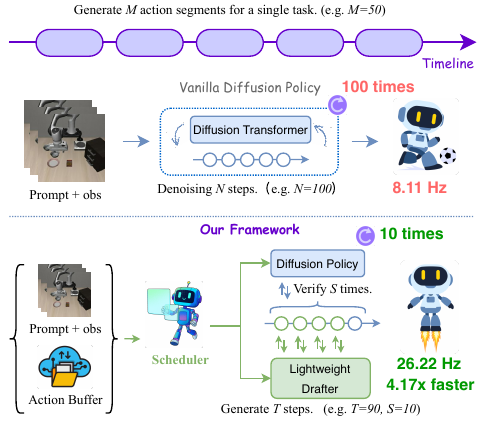}
    \caption{
        \textbf{Vanilla Diffusion Policy (DP)  vs.  Our \sysname.}
        For embodied tasks, DP completes a task through multiple interactions with the environment, generating several segments of actions.
        Each segment typically requires hundreds of model calls for denoising, making the process highly time-consuming.
        \sysname introduces a customized speculative decoding framework for temporal-complexity-aware accelerating DP.
        A lightweight drafter generates multiple denoising results, which are verified in parallel by the DP for near-lossless acceleration.
        A scheduler further adjusts key speculative decoding parameters—such as the number of drafts and the acceptance rate—based on the complexity of different task phases, achieving adaptive and efficient acceleration.
    }
    \label{fig:overview}
\end{figure}

Recent research has increasingly focused on accelerating diffusion models (DMs) to enable faster inference. 
General model compression techniques, including pruning and quantization~\cite{q-dit,wan2025pruning,zhang2024laptop,li2023q}, reduce computational cost by simplifying the diffusion backbone and have shown effectiveness in static generation tasks. 
Building on these advances, a growing body of work explores specific acceleration techniques for DP, where cache-based methods~\cite{Ji2025BlockwiseAC,yang2025efficientvla} reuse intermediate features from previous denoising steps to eliminate redundant computation during action generation.
While effective to some extent, these approaches inevitably alter internal feature representations, resulting in \textit{lossy acceleration} and constrained overall speedup.
Furthermore, they remain largely static and fail to account for the temporal correlations and evolving task complexity inherent in embodied control.

Meanwhile, speculative decoding has emerged as a lossless and complexity-adaptive acceleration paradigm, where a lightweight drafter proposes multiple candidates that are verified in parallel by a stronger model. 
This strategy has achieved notable success in LLMs \cite{medusa,eagle,eagle-2,eagle-3}, delivering substantial speedups without compromising fidelity, and has also been adapted to DMs by accelerating the denoising process \cite{Bortoli2025AcceleratedDM,Hu2025DiffusionMA}.
However, these approaches primarily target compute-bound DMs for image and video generation, where GPU computation dominates and the verification stage offsets most potential savings, resulting in limited acceleration. 
In contrast, \textbf{DP is inherently I/O-bound} as it generates low-dimensional action vectors instead of high-resolution visual data. 
This property makes speculative decoding particularly well-suited for DP inference. 
Yet DP differs fundamentally from conventional DMs: it must produce \textbf{sequential} action segments while continuously interacting with an evolving environment, where task difficulty and dynamics change over time, making temporal adaptivity essential for stable and efficient acceleration. 
As a result, temporal adaptivity becomes crucial for maintaining both stability and efficiency.
This motivates a temporal-complexity-aware speculative decoding framework that preserving behavioral consistency and robust control performance.

Therefore, we propose \sysname, the first temporal-complexity-aware speculative decoding framework designed to accelerate DP, as illustrated in Fig.~\ref{fig:overview}.
This presents two key challenges:
(1) how to match the DP’s denoising quality at lower cost under time-varying task difficulty, and
(2) how to dynamically adjust computation based on task difficulty to ensure stable and efficient control.

For the first challenge, prior studies leverage either the differences between consecutive time steps as drafts or the exchangeability of trajectories to enable self-speculation and parallel verification.
However, embodied tasks are highly diverse, and their multi-segment action sequences often exhibit significant distributional differences, making these approaches difficult to generalize and largely limited to image generation.
To address this, we adopt a single Transformer block as the drafter and train it via knowledge distillation, where DP serves as the teacher and the drafter is optimized to minimize both its deviation from the ground truth and its discrepancy from the teacher’s outputs.
The drafter shares the same input–output interface as DP, performs multiple denoising steps in its place, and has its results verified in parallel by DP.
During each speculative decoding round, accepted denoising steps are retained, while the first rejected one is corrected via Reflection-Maximal Coupling to better align with DP’s denoising distribution.
This process repeats iteratively until the denoising is complete.

For the second challenge, the temporal characteristics of embodied tasks cause task difficulty to vary dynamically over time. 
As shown in Fig.~\ref{fig:400step}, empirical analysis reveals an inverse relationship between end-effector velocity and the number of accepted drafts—faster, coarse-grained motions lead to fewer accepted drafts, while slower, fine-grained actions yield more. 
This indicates that using fixed parameters for speculative decoding is suboptimal. 
Existing approaches, such as SP-VLA \cite{li2025sp}, distinguish task difficulty by velocity, which limits their generalization across diverse behaviors. 
To address this, we introduce a Reinforcement Learning (RL)–based scheduler that dynamically adjusts speculative decoding parameters according to task phase, enabling adaptive and efficient acceleration.
Extensive experiments demonstrate that \sysname achieves a draft acceptance rate of \textbf{over 85\%}, increasing the inference frequency from \textbf{8 Hz} to \textbf{26 Hz} while reducing the Number of Function Evaluations (NFE) by 4.17$\times$, all while maintaining comparable task performance.
Real-world results further demonstrate up to a \textbf{3.8$\times$} NFE reduction and improved success rates on real-time tasks (e.g., a \textbf{23\%} increase on the conveyor-belt task), confirming its effectiveness for real-time control.
The main contributions are as follows:
\begin{itemize}
    \item[(1)] We propose \sysname, a temporal-complexity-aware speculative decoding framework for DP, enabling frequency-adaptive and accuracy-preserving acceleration. To the best of our knowledge, this is the first work that brings speculative decoding to DP with temporal adaptivity.
    \item[(2)] We propose a speculative decoding framework tailored for DP. To match the base model’s denoising quality at lower cost under time-varying task difficulty, we distill a Transformer-based drafter to imitate the base model and replace its costly multi-step denoising calls, while DP verifies its outputs in parallel.
    \item[(3)] We propose a temporal-complexity-aware speculative decoding scheduler. To dynamically adjust computation based on task difficulty, we train a PPO-based scheduler that formulates DP execution as a Markov process and adaptively tunes speculative parameters over time.
\end{itemize}

\section{Related Work}
\label{sec:related_work}

\subsection{Diffusion Policy Acceleration}
Benefiting from the strong modeling capacity and widespread success of DMs \cite{weng2024dcm,jiang2024animediff,xu2024sgdm}, DP has emerged as a leading framework for visuomotor policy learning. 
It leverages the generative capability of DMs to produce continuous and multimodal robotic actions conditioned on visual observations and language instructions~\cite{dp,MartinezCantin2007ActivePL}.
Despite their strong performance, DP suffer from high inference latency caused by iterative denoising, which limits their applicability in real-time, high-frequency robotic control.

To tackle this challenge, extensive research has focused on accelerating diffusion-based generative models and their extensions to embodied policies~\cite{li2025sp,11146899,pei2025action,yu2025survey,guan2025efficient,li2023ddpg,liu2024novel}. 
A variety of techniques are  applied to accelerate diffusion models in the field of image and video generation, such as quantization \cite{q-dit,li2023q,q-1,q-2,q-3,q-4,q-5}, feature cache \cite{Zhou2024VariationalDO, teacache, speca, ca-1,ca-2,ca-3}.
In contrast, EfficientVLA~\cite{yang2025efficientvla} and BAC~\cite{Ji2025BlockwiseAC} are specifically designed for DP acceleration, where the latter introduces block-wise adaptive caching to selectively refresh upstream features and mitigate error propagation. 
However, most existing methods modify intermediate representations during inference, resulting in lossy acceleration, and few explicitly account for the temporal redundancy inherent in DP.

\subsection{Speculative Decoding}
Speculative decoding has recently emerged as a principled paradigm for accelerating autoregressive generation. 
Originally developed for LLMs~\cite{sp-1,sp-2,sp-3}, it employs a lightweight \textit{drafter} to propose multiple future tokens that are verified in parallel by a stronger \textit{target} model within a single forward pass, achieving provable lossless acceleration. 
Representative approaches such as the Medusa and Eagle series~\cite{medusa,eagle,eagle-2,eagle-3} further enhance efficiency through multi-head drafting and hierarchical verification mechanisms.

Several studies have extended speculative decoding to continuous generative processes. 
\cite{Bortoli2025AcceleratedDM} introduced the frozen target draft model, which accelerates diffusion denoising by using stepwise differences as drafts and applies reflection-maximal coupling to preserve the original sampling distribution.
Building on this, \cite{Hu2025DiffusionMA} proposed autospeculative decoding, eliminating the need for a separate drafter by exploiting trajectory exchangeability for self-speculation and parallel verification.
While these methods achieve lossless acceleration in DDPMs and LDMs, they are primarily designed for compute-bound diffusion models in image and video generation, and the verification overhead largely cancels out the benefit—resulting in the counterintuitive phenomenon that even using 8 GPUs for verification yields only about a 1.8$\times$ speedup.

In contrast, DP is inherently I/O-bound and therefore better suited for speculative decoding. 
The drafter operates at a much lower computational cost than the base policy, allowing parallel verification to be substantially more efficient than in compute-bound DMs. 
However, unlike standard DMs, DP must generate sequential action segments through interactions with the environment, where early prediction errors can accumulate and degrade overall task performance. 
To fix them, we introduce \textbf{\sysname}, a temporal-complexity-aware speculative decoding framework that enables adaptive, near-lossless acceleration for embodied tasks.

\section{Method}
\label{sec:method}

\begin{figure*}[t]
    \centering
    \includegraphics[scale=1.05]{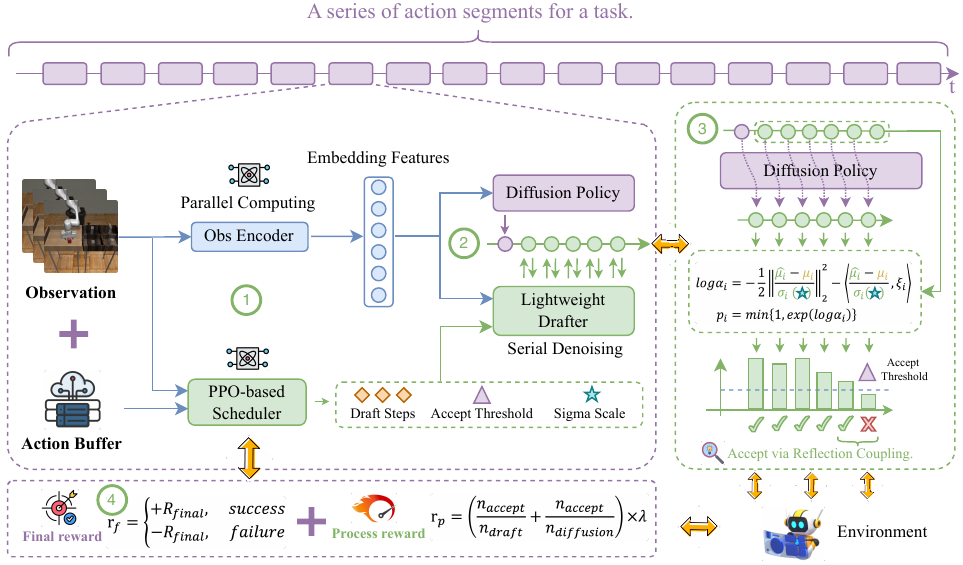}
    \caption{
        \textbf{The framework of \sysname.}
        \sysname is the first temporal-aware speculative decoding framework designed to accelerate Diffusion Policy.
        \textcircled{1} \textbf{Decision stage:} A PPO-based scheduler evaluates time-varying task difficulty using past observations and actions, and produces adaptive speculative parameters. It operates in parallel with the observation encoder, adding no extra inference latency.
        \textcircled{2} \textbf{Denoising stage:} Guided by the scheduler, a lightweight drafter executes multiple denoising steps sequentially, effectively replacing expensive DP calls and reducing computational load.
        \textcircled{3} \textbf{Verification stage:} DP verifies all drafted steps in parallel, accepting those that pass validation and correcting the first rejected draft via reflection-maximal coupling to better align it with the target distribution.
        \textcircled{4} \textbf{Reward generation:} Based on task progress, the framework provides process rewards that encourage the scheduler to produce more valid drafts, and a final success-driven reward upon task completion, jointly guiding the scheduler toward efficient and successful task execution.
    }   
    \label{fig:framework}
\end{figure*}

\subsection{Preliminaries}
\textbf{Speculative decoding.} 
Speculative decoding leverages a lightweight draft model $p$ to generate a short trajectory of length $L$ in a single forward pass, while the large-scale target model $q$ verifies the generated drafts in parallel.
This approach reduces the number of target model invocations, thereby improving inference efficiency.
The joint distribution of the drafted latents is formulated as:
\begin{equation}
    \tilde{\mathbf{x}}_{t+1:t+L} \sim 
    p\!\left(\tilde{\mathbf{x}}_{t+1:t+L} \mid \mathbf{x}_t\right)
    =
    \prod_{j=1}^{L}
    p\!\left(
        \tilde{\mathbf{x}}_{t+j}
        \mid
        \tilde{\mathbf{x}}_{t+1:t+j-1},\,
        \mathbf{x}_t
    \right),
\end{equation}
where $\mathbf{x}_t$ denotes the current diffusion latent at timestep $t$,  
and $\tilde{\mathbf{x}}_{t+j}$ denotes the $j$-th drafted latent.
Each drafted latent is accepted according to the standard likelihood-ratio test:
\begin{equation}
    \alpha_j(\tilde{\mathbf{x}}_{t+j})
    =
    \min\!\left(
        1,\;
        \frac{
            q\!\left(
                \tilde{\mathbf{x}}_{t+j}
                \mid
                \tilde{\mathbf{x}}_{t+1:t+j-1},\,
                \mathbf{x}_t
            \right)
        }{
            p\!\left(
                \tilde{\mathbf{x}}_{t+j}
                \mid
                \tilde{\mathbf{x}}_{t+1:t+j-1},\,
                \mathbf{x}_t
            \right)
        }
    \right).
\end{equation}
The first rejection index is defined as:
\begin{equation}
    j^\star
    =
    \min\!\left\{
        j \in \{1,\dots,L\}
        \;:\;
        U_j >
        \alpha_j\!\left(\tilde{\mathbf{x}}_{t+j}\right)
    \right\},
\end{equation}
where $U_j \sim \mathrm{Unif}(0,1)$ are independent uniform random variables.
All drafted latents after $j^\star$ are discarded, and a corrective coupling is applied at step $t + j^\star - 1$ to obtain a latent consistent with the target distribution.

The window length $L$ and the closeness of $p$ to $q$ jointly control the trade-off between parallelism and the probability of encountering a rejection within the window.

\noindent \textbf{Reflection-maximal coupling.} 
For continuous, approximately Gaussian one-step conditionals, we apply \textit{reflection-maximal coupling} to correct the first rejected draft sample through a single deterministic mapping.
Let
\begin{equation}
    r(\mathbf{x}) = \mathcal{N}(\mathbf{x};\, \mathbf{m}_r, \boldsymbol{\Sigma}),
    \qquad
    s(\mathbf{x}) = \mathcal{N}(\mathbf{x};\, \mathbf{m}_s, \boldsymbol{\Sigma}),
    \label{eq:reflection_1}
\end{equation}
where $r(\mathbf{x})$ and $s(\mathbf{x})$ denote the draft and target Gaussian densities with means $\mathbf{m}_r$ and $\mathbf{m}_s$ and shared covariance $\boldsymbol{\Sigma}$.
Draw $\tilde{\mathbf{x}} \sim r$ and perform the acceptance test:
\begin{equation}
    A
    =
    \mathbf{1}\!\left(
        U \le
        \min\!\left(
            1,\;
            \frac{s(\tilde{\mathbf{x}})}{r(\tilde{\mathbf{x}})}
        \right)
    \right),
    \qquad
    U \sim \mathrm{Unif}(0,1).
    \label{eq:reflection_2}
\end{equation}
If $A = 1$, accept $\mathbf{x} = \tilde{\mathbf{x}}$. 
If $A = 0$, obtain a corrected sample by reflecting $\tilde{\mathbf{x}}$ across the hyperplane orthogonal to $\boldsymbol{\Delta} := \mathbf{m}_r - \mathbf{m}_s$.
In the isotropic case $\boldsymbol{\Sigma} = \sigma^2 \mathbf{I}$, the reflection takes the form:
\begin{equation}
    \mathbf{x}
    =
    \mathbf{m}_s
    +
    \big(\mathbf{I} - 2 \mathbf{e} \mathbf{e}^\top\big)
    (\tilde{\mathbf{x}} - \mathbf{m}_r),
    \qquad
    \mathbf{e} := 
    \frac{\boldsymbol{\Delta}}{\lVert \boldsymbol{\Delta} \rVert_2}.
    \label{eq:reflection_3}
\end{equation}

Eqs.~\eqref{eq:reflection_1}--\eqref{eq:reflection_3} depend only on the means $\mathbf{m}_r$, $\mathbf{m}_s$, the shared covariance $\boldsymbol{\Sigma}$, and a single draw $\tilde{\mathbf{x}} \sim r$.  
The mapping in Eq.~\eqref{eq:reflection_3} generates a sample whose marginal distribution follows $s$, while minimizing the discrepancy between the rejected draft and its corrected counterpart.

\subsection{Speculative Drafting and Verification Framework}
\noindent \textbf{Draft Model Training.} 
The draft model is a lightweight approximator of the target model, designed for fast speculative rollouts while maintaining behavioral alignment. 
Since a lightweight architecture is sufficient for approximate denoising, we employ a single-layer Transformer block as the draft model, which shares the same encoder and DDPM or DDIM scheduler with the target model $M_\phi$.

Specifically, we train the draft model $\hat{M}_\theta$ via knowledge distillation with two objectives. 
(1) The first component is a prediction-level distillation loss that aligns the draft model output $\hat{\mathbf{m}}_t^\theta$ with the target model output $\mathbf{m}_t^\phi$:
\begin{equation}
    \mathcal{L}_{\mathrm{pred}}(\theta)
    =
    \mathbb{E}_{t, \mathbf{x}_0, \boldsymbol{\epsilon}}
    \left\|
        \hat{\mathbf{m}}_t^\theta - \mathbf{m}_t^\phi
    \right\|_2^2,
\end{equation}
where $t$ is the diffusion timestep, $\mathbf{x}_0$ is the clean action, and $\boldsymbol{\epsilon}$ is the Gaussian noise used to construct the noisy latent.
(2) The second component is a scheduler-aware normalized loss that aligns the draft model with the data-consistent denoising behavior of the target model:
\begin{equation}
    \mathcal{L}_{\mathrm{norm}}(\theta)
    =
    \mathbb{E}_{t, \mathbf{x}_0, \boldsymbol{\epsilon}}
    \left\|
        \frac{
            \hat{\boldsymbol{\mu}}_t^\theta - \boldsymbol{\mu}_t^\phi
        }{
            \sigma_t
        }
    \right\|_2^2,
\end{equation}
where $\boldsymbol{\mu}_t^\phi$ and $\hat{\boldsymbol{\mu}}_t^\theta$ are the data-aligned DDPM means computed from the target and draft predictions, and $\sigma_t$ is the DDPM standard deviation at timestep $t$.

The full objective combines both terms:
\begin{equation}
    \mathcal{L}(\theta)
    =
    \lambda_{1} \, \mathcal{L}_{\mathrm{pred}}(\theta)
    +
    \lambda_{2} \, \mathcal{L}_{\mathrm{norm}}(\theta),
\end{equation}
with $\lambda_{1}, \lambda_{2} > 0$.
The target model parameters $\boldsymbol{\phi}$ are kept frozen throughout training,
while only the draft parameters $\boldsymbol{\theta}$ are updated.

\noindent \textbf{Draft Generation Procedure.}
During the inference phase, speculative drafts are generated through a short and efficient forward rollout using the lightweight draft model, conditioned on the current denoised estimate produced by the target model.
Let $M$ denote the target model and $\mathcal{S}$ the DDPM scheduler.
At each time step $t$, the target model predicts the noise estimate
$\mathbf{m}_t = M(\mathbf{x}_t; t)$,
and the scheduler $\mathcal{S}$ produces the immediate one-step denoised candidate
$\mathbf{x}_{t-1}^{(0)} = \mathcal{S}\!\left(\mathbf{m}_t,\, t,\, \mathbf{x}_t\right)$.

Starting from $\mathbf{x}_{t-1}^{(0)}$, a lightweight draft model $\hat{M}$ performs an autoregressive-style multi-step rollout for up to $K$ steps.
For each draft step $k = 1, \dots, K$, we compute the draft prediction at time $t - k$ and apply the same DDPM scheduler to obtain the next draft sample:
\begin{equation}
\begin{aligned}
    \hat{\mathbf{m}}_{t-k} &= \hat{M}\!\left(\mathbf{x}_{t-k+1}^{(\mathrm{draft})};\, t - k\right), \\
    \mathbf{x}_{t-k}^{(\mathrm{draft})} &= \mathcal{S}\!\left(\hat{\mathbf{m}}_{t-k},\, t - k,\, \mathbf{x}_{t-k+1}^{(\mathrm{draft})}\right).
\end{aligned}
\end{equation}

This rollout continues until either the maximum draft horizon $K$ is reached or the diffusion process arrives at $t = 0$. 

\begin{table*}[t]
    \caption{\textbf{Benchmark on Proficient Human (PH) demonstration data.}}
    \centering
    \small
    \setlength{\tabcolsep}{3pt}
    \begin{threeparttable}
    \scalebox{0.99}{%
    \begin{tabular*}{\linewidth}{@{\extracolsep{\fill}}lccccccccc@{}}
        \toprule
        \multirow{2}{*}{\textbf{Method}} &
        \multicolumn{6}{c}{\textbf{Success Rate (\%, $\uparrow$)}} &
        \multirow{2}{*}{\textbf{\makecell{AVG \\ (\%, $\uparrow$)}}} &
        \multirow{2}{*}{\textbf{\makecell{NFE \\ (\%, $\downarrow$)}}} &
        \multirow{2}{*}{\textbf{Speed $\times$}} \\
        \cmidrule(lr){2-7}
        & Lift & Can & Square & Transport & Tool & Push-T &  &  &  \\
        \midrule
        Diffusion Policy \cite{dp}
            & 100 / 100 & 95 / 97 & 82 / 88 & 78 / 81 & 43 / 53 & 59 / 64
            & 76 / 80 & 100 / 100 & -- \\
        \midrule
        Frozen Target Draft \cite{Bortoli2025AcceleratedDM}
            & 90 / 92 & 82 / 84 & 70 / 70 & 50 / 64 & 26 / 34 & 16 / 17
            & 56 / 60 & 33 / 32 & 3.03 / 3.13 \\
        SpeCa \cite{speca} 
            & \textbf{100 / 100} & 92 / 96 & \textbf{88} / 88 & 68 / 72 & 38 / 40 & 67 / \textbf{69}
            & 76 / 78 & 34 / 35 & 2.94 / 2.86 \\
        BAC \cite{Ji2025BlockwiseAC}
            & \textbf{100 / 100} & 94 / \textbf{97} & 82 / \textbf{89} & 77 / 82 & 49 / \textbf{55} & 59 / 62
            & 77 / \textbf{81} & -- / -- & 3.40 / 3.40 \\
        \midrule
        \sysname
            & \textbf{100 / 100}
            & \textbf{95} / 95
            & 86 / 88
            & \textbf{87 / 84}
            & \textbf{61} / 52
            & \textbf{79} / 63
            & \textcolor{red}{\textbf{85} / 80}
            & \textcolor{blue}{\textbf{24 / 24}}
            & \textcolor{blue}{\textbf{4.17 / 4.17}} \\
        \bottomrule
    \end{tabular*}%
    }
    \begin{tablenotes}
        \footnotesize
        \item Accuracy, NFE, and acceleration performance are reported; \sysname achieves a 4.17$\times$ acceleration on such tasks while also improving accuracy by 9\%.
        On certain tasks, such as Transport and Tool, \sysname even surpasses the baseline. 
        This improvement can be attributed to the implicit filtering effect of \sysname, which enhances the smoothness and stability of the manipulator during task execution.
        Consistent with the original DP paper \cite{dp}, each cell reports results obtained from different checkpoints.
    \end{tablenotes}
    \end{threeparttable}
    \label{tab:main_result_ph}
\end{table*}
\begin{table*}[!htbp]
    \caption{\textbf{Benchmark on Mixed Human (MH) demonstration data.}}
    \centering
    \small
    \setlength{\tabcolsep}{3pt}
    \begin{threeparttable}
    \scalebox{0.99}{%
    \begin{tabular*}{\linewidth}{@{\extracolsep{\fill}}lccccccc@{}}
        \toprule
        \multirow{2}{*}{\textbf{Method}} &
        \multicolumn{4}{c}{\textbf{Success Rate (\%, $\uparrow$)}} &
        \multirow{2}{*}{\textbf{\makecell{AVG \\ (\%, $\uparrow$)}}} &
        \multirow{2}{*}{\textbf{\makecell{NFE \\ (\%, $\downarrow$)}}} &
        \multirow{2}{*}{\textbf{Speed $\times$}} \\
        \cmidrule(lr){2-5}
        & Lift & Can & Square & Transport &  &  &  \\
        \midrule
        Diffusion Policy \cite{dp}
            & 99 / 100 & 92 / 97 & 76 / 79 & 35 / 46
            & 76 / 81 & 100 / 100 & -- \\
        \midrule
        Frozen Target Draft \cite{Bortoli2025AcceleratedDM}
            & 66 / 62 & 60 / 54 & 50 / 62 & 20 / 42
            & 49 / 55 & 34 / 33 & 2.94 / 3.03 \\
        SpeCa \cite{speca}
            & 96 / \textbf{100} & 86 / 90 & 76 / 68 & \textbf{34} / 40
            & 73 / 75 & 35 / 37 & 2.86 / 2.70 \\
        BAC \cite{Ji2025BlockwiseAC}
            & \textbf{99} / 98 & \textbf{95} / \textbf{97} & \textbf{77} / \textbf{79} & 30 / 46
            & \textbf{75} / 80 & -- / -- & 3.41 / 3.41 \\
        \midrule
        \sysname
            & \textbf{99} / 99 & 92 / \textbf{97} & 74 / 78 & \textbf{34} / \textbf{62}
            & \textcolor{red}{\textbf{75 / 84}} & \textcolor{blue}{\textbf{26 / 26}} & \textcolor{blue}{\textbf{3.84 / 3.84}} \\
        \bottomrule
    \end{tabular*}%
    }
    \begin{tablenotes}
        \footnotesize
        \item Accuracy, NFE, and acceleration performance are reported; \sysname maintains comparable empirical accuracy on this task while delivering a 3.84$\times$ acceleration.
        Due to the inherent complexity of this type of data, the training of drafter is affected to some extent; nevertheless, we still achieve near-lossless performance.
    \end{tablenotes}
    \end{threeparttable}
    \label{tab:main_result_mh}
\end{table*}
\begin{table*}[!htbp]
    \caption{\textbf{Benchmark on multi-stage task (Kitchen \& BP).}}
    \centering
    \small
    \setlength{\tabcolsep}{3pt}
    \begin{threeparttable}
    \scalebox{0.99}{%
    \begin{tabular*}{\linewidth}{@{\extracolsep{\fill}}lccccccccc@{}}
        \toprule
        \multirow{2}{*}{\textbf{Method}} &
        \multicolumn{6}{c}{\textbf{Success Rate (\%, $\uparrow$)}} &
        \multirow{2}{*}{\textbf{\makecell{AVG \\ (\%, $\uparrow$)}}} &
        \multirow{2}{*}{\textbf{\makecell{NFE \\ (\%, $\downarrow$)}}} &
        \multirow{2}{*}{\textbf{Speed $\times$}} \\
        \cmidrule(lr){2-7}
        & BP$_{p1}$ & BP$_{p2}$ & Kit$_{p1}$ & Kit$_{p2}$ & Kit$_{p3}$ & Kit$_{p4}$ &  &  &  \\
        \midrule
        Diffusion Policy \cite{dp}
            & 98 / 98 & 98 / 96 & 100 / 100 & 100 / 100 & 100 / 100 & 100 / 100
            & 99 / 99 & 100 / 100 & -- \\
        \midrule
        Frozen Target Draft \cite{Bortoli2025AcceleratedDM}
            & 85 / 86 & 1 / 2 & \textbf{100 / 100} & \textbf{100 / 100} & \textbf{100 / 100} & \textbf{100 / 100}
            & 81 / 81 & 33 / 33 & 3.03 / 3.03 \\
        SpeCa \cite{speca}
            & 97 / 97 & 85 / 85 & \textbf{100 / 100} & \textbf{100 / 100} & \textbf{100 / 100} & \textbf{100 / 100}
            & 97 / 97 & 37 / 37 & 2.70 / 2.70 \\
        BAC \cite{Ji2025BlockwiseAC}
            & \textbf{100} / 99 & \textbf{97} / 95 & \textbf{100} / 99 & \textbf{100} / 99 & \textbf{100} / 99 & 94 / 97
            & \textbf{99} / 98 & -- / -- & 3.60 / 3.60 \\
        \midrule
        \sysname
            & 99 / \textbf{100} & \textbf{97 / 97} & \textbf{100 / 100} & \textbf{100 / 100}
            & \textbf{100 / 100} & \textbf{100 / 100}
            & \textcolor{red}{\textbf{99 / 99}} & \textcolor{blue}{\textbf{27 / 27}} & \textcolor{blue}{\textbf{3.70 / 3.70}} \\
        \bottomrule
    \end{tabular*}%
    }
    \begin{tablenotes}
        \footnotesize
        \item In the Kitchen task, $p_x$ denotes the frequency of interacting with $x$ or more objects (e.g., microwave, burners, oven switch, kettle); in the BP task, $p_x$ represents the success rate under progressively harder control phases; \sysname{} maintains comparable empirical accuracy on this task while delivering a 3.7$\times$ acceleration.
    \end{tablenotes}
    \end{threeparttable}
    \label{tab:main_result_3}
\end{table*}

\begin{figure*}[t]
    \centering
    \includegraphics[width=\linewidth]{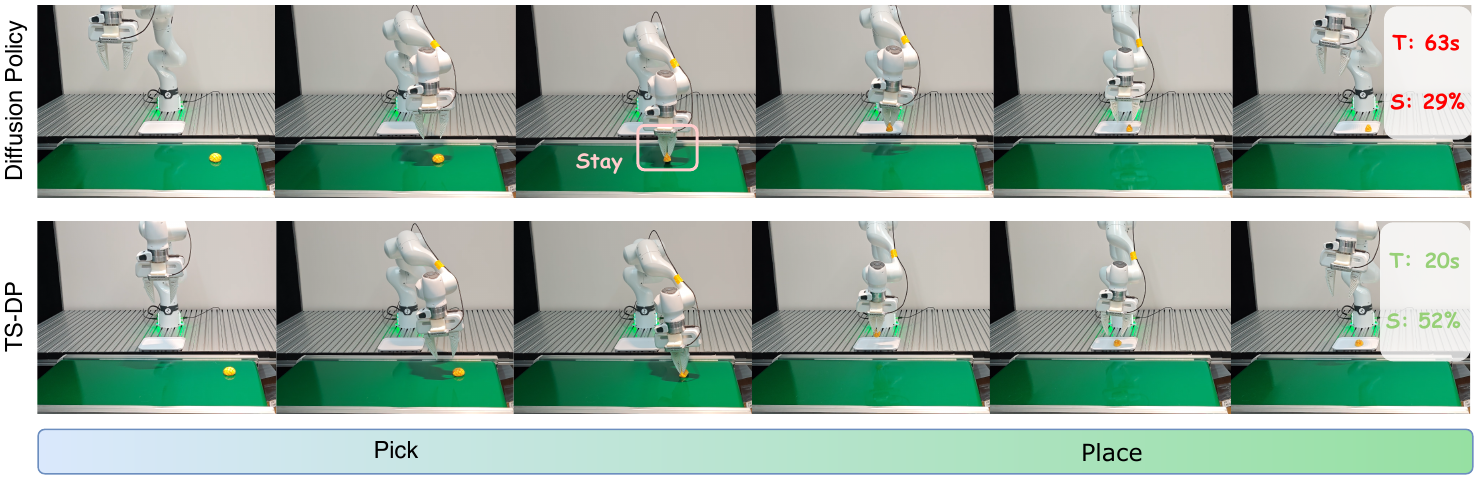}
    \caption{
        \textbf{Conveyor-Belt Pick-and-Place Task.}
        This figure shows a Franka robotic arm picking pineapple buns from a moving conveyor belt and placing them onto a plate. \sysname finishes in 20s, compared to 63s for DP, yielding a 3.2$\times$ speedup. This improved real-time performance also increases the success rate from 29\% to 52\%.
        Here, \textit{T} denotes time in seconds (s), and \textit{S} denotes success rate in percentage (\%).
        }
    \label{fig:real_infer}
\end{figure*}

\begin{figure*}[t]
    \centering
    \includegraphics[width=\linewidth]{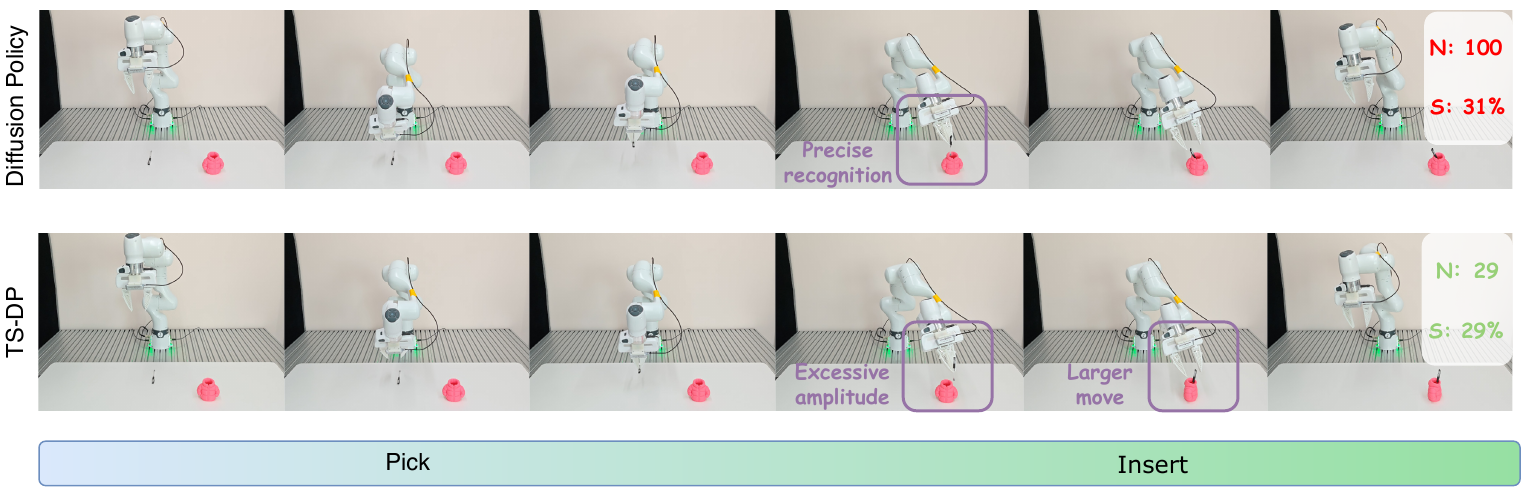}
    \caption{
        \textbf{Pen-to-Holder Insertion Task.}
        This task is designed to examine the impact of \sysname on high-difficulty, high-precision manipulation. 
        As shown in the figure, DP exhibits greater precision at critical moments, such as more accurate alignment with the pen holder, resulting in smaller displacements. 
        Although \sysname shows minor inaccuracies in these cases, it is able to quickly correct them and successfully complete the task.
        It ultimately maintains a 3.4$\times$ NFE speedup, with the success rate decreasing by only 2\%.
        Here, \textit{N} denotes NFE, and \textit{S} denotes success rate(\%).
        }
    \label{fig:pick_and_insert}
\end{figure*}

\begin{figure*}[t]
    \centering
    \includegraphics[width=\linewidth]{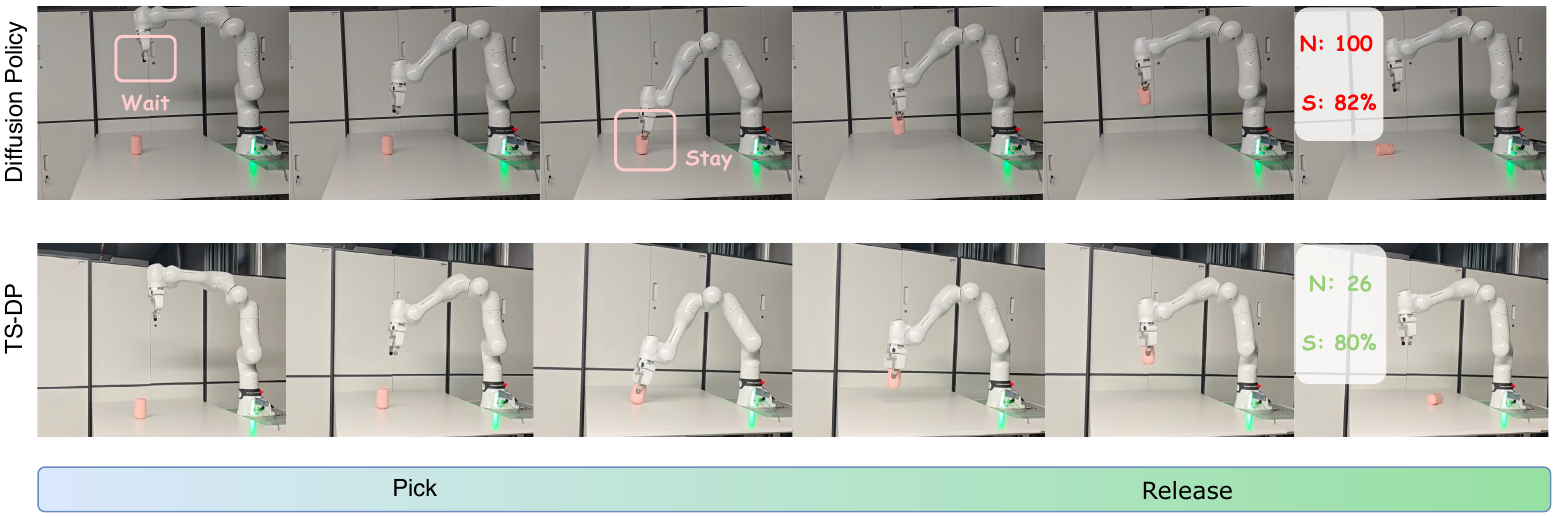}
    \caption{
        \textbf{Pick-and-Release Task.}
        As shown in the figure, DP exhibits prolonged pauses both at the start and after grasping, whereas \sysname effectively mitigates this issue, achieving a 3.8$\times$ NFE speedup with only a 2\% accuracy drop.
        The task closely matches the difficulty of the official simulation and real-world benchmarks, providing a clear setting to showcase the acceleration achieved by our method.
        }
    \label{fig:pick_and_release}
\end{figure*}

\begin{figure}[t]
    \centering
    \subfloat[Smoothed variance\label{fig:smooth_variance}]{
        \begin{minipage}[b]{0.47\linewidth}
            \centering
            \includegraphics[width=\linewidth]{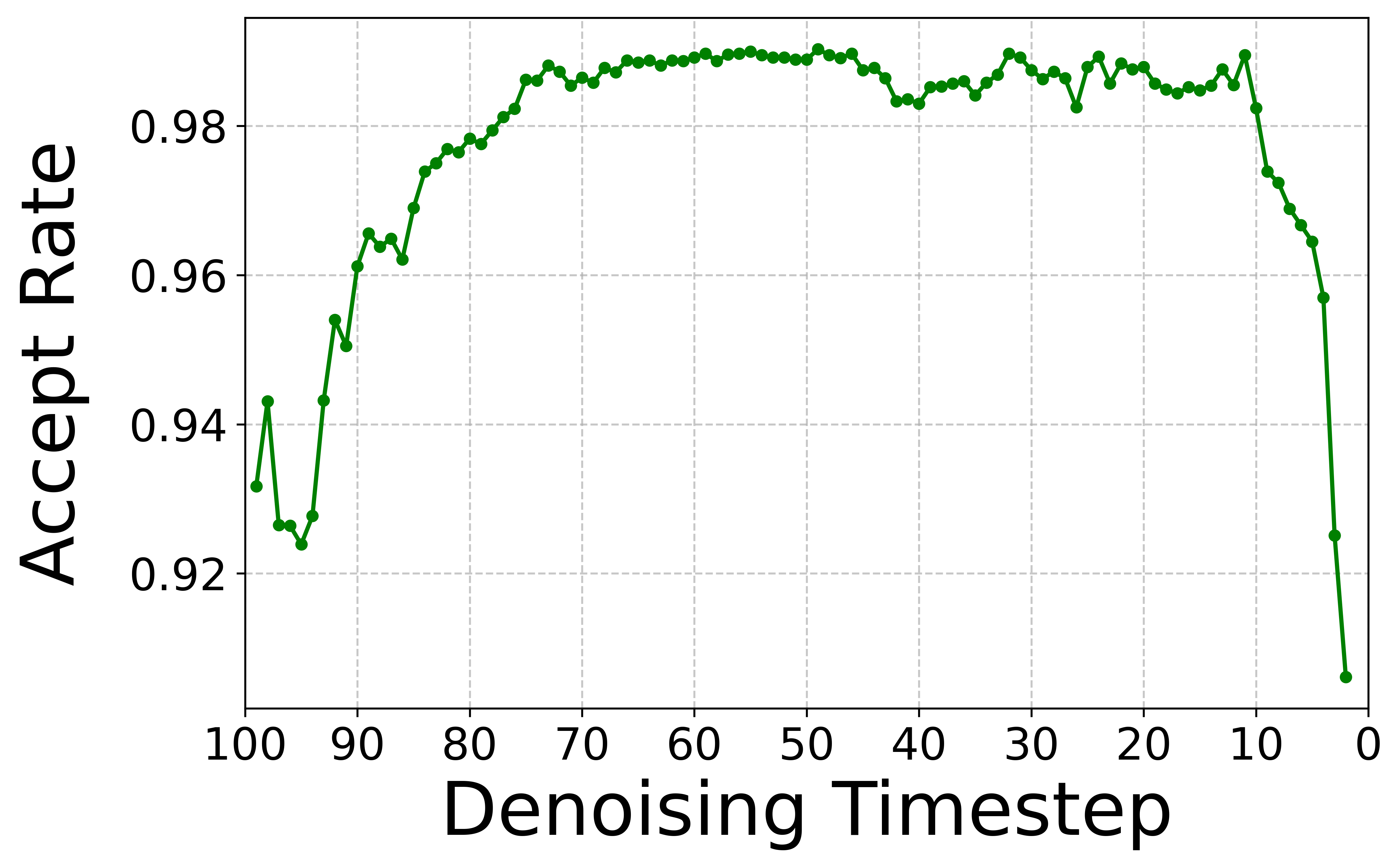}
        \end{minipage}
    }
    \hfill
    \subfloat[Standard variance\label{fig:standard_variance}]{
        \begin{minipage}[b]{0.47\linewidth}
            \centering
            \includegraphics[width=\linewidth]{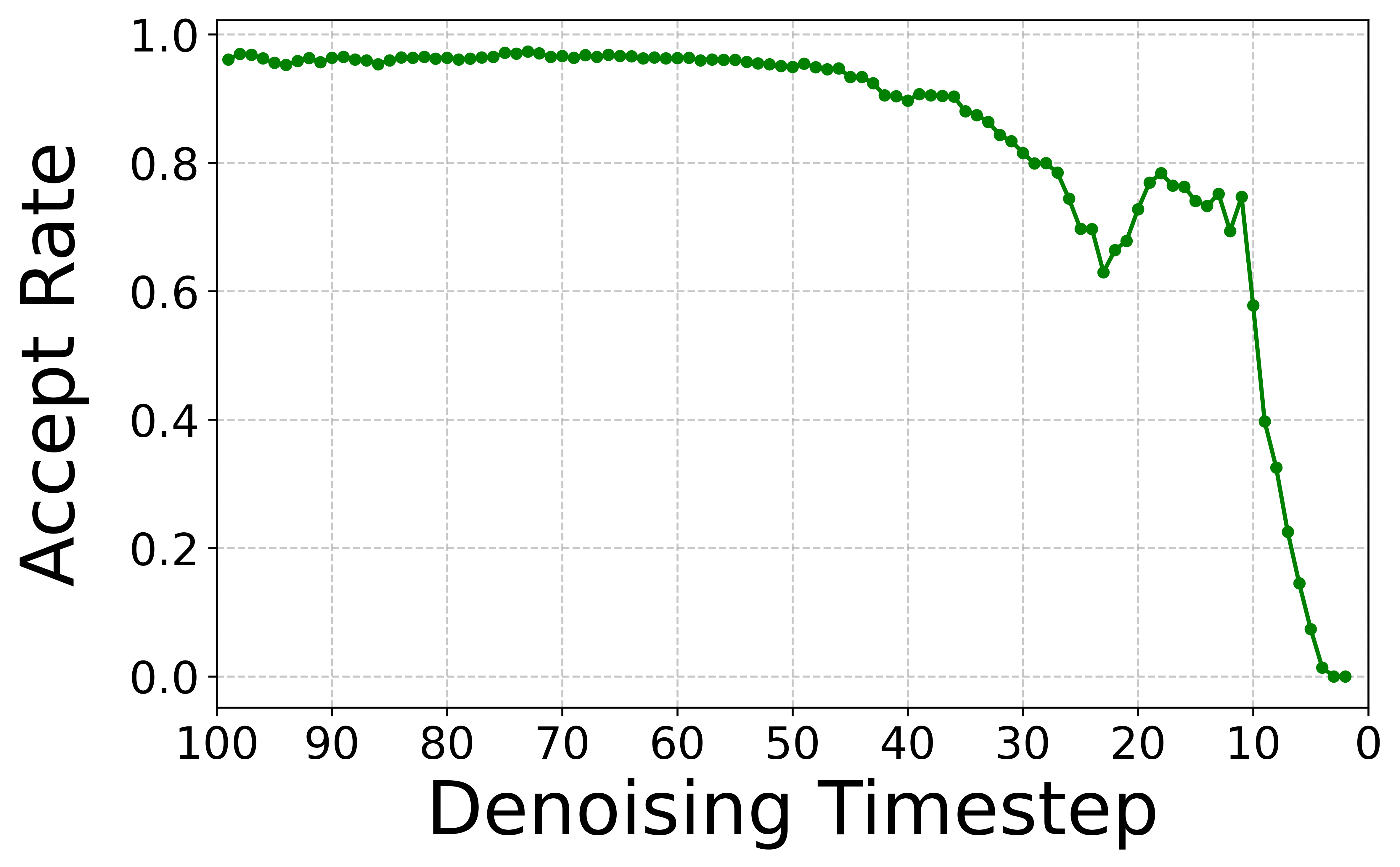}
        \end{minipage}
    }
    \caption{\textbf{Impact of Drafter Parameters on Draft Acceptance Rate.} Evolution of the Metropolis--Hastings acceptance probability throughout the 100-step denoising process.}
    \label{fig:variance_comparison}
\end{figure}

As shown in Fig.~\ref{fig:smooth_variance}, the acceptance probability displays clear phase-dependent trends: it is low during both the early high-noise and late low-noise stages, while remaining high and stable throughout the intermediate denoising phase.
To balance acceleration and sample fidelity, \sysname treats the draft horizon $K$ as a timestep-dependent parameter, using smaller $K$ in high- and low-noise stages and larger $K$ in intermediate phases where predictions are more reliable. 
The scheduler dynamically adjusts $K$ to maintain high draft acceptance and maximize speedup without degrading perceptual quality.

\noindent \textbf{Verification of Speculative Drafts.} 
Given the one-step candidate $\mathbf{x}_{t-1}^{(0)}$ from the target model and a sequence of $K$ draft proposals $\{\mathbf{x}_{t-1}^{(1)},\, \mathbf{x}_{t-2}^{(2)},\, \dots,\, \mathbf{x}_{t-K}^{(K)}\}$ generated by the draft model, we form the candidate set $\mathcal{C}$.
Verification is performed by evaluating the frozen target model $M_{\phi}$ on all elements of $\mathcal{C}$ in a single batched forward pass.
For each candidate, we obtain the conditional mean $\boldsymbol{\mu}_i$ and the effective standard deviation $\sigma_i$ from the scheduler.
As shown in Fig.~\ref{fig:standard_variance}, the standard deviation is critical for draft acceptance: without adjustment, acceptance collapses in later denoising stages due to numerical overconfidence. 
We therefore treat $\sigma_i$ as a tunable parameter and adapt it via the scheduler to maintain stable acceptance and consistent speculative acceleration across the trajectory.

\begin{figure}[t]
    \centering
    \includegraphics[width=\linewidth]{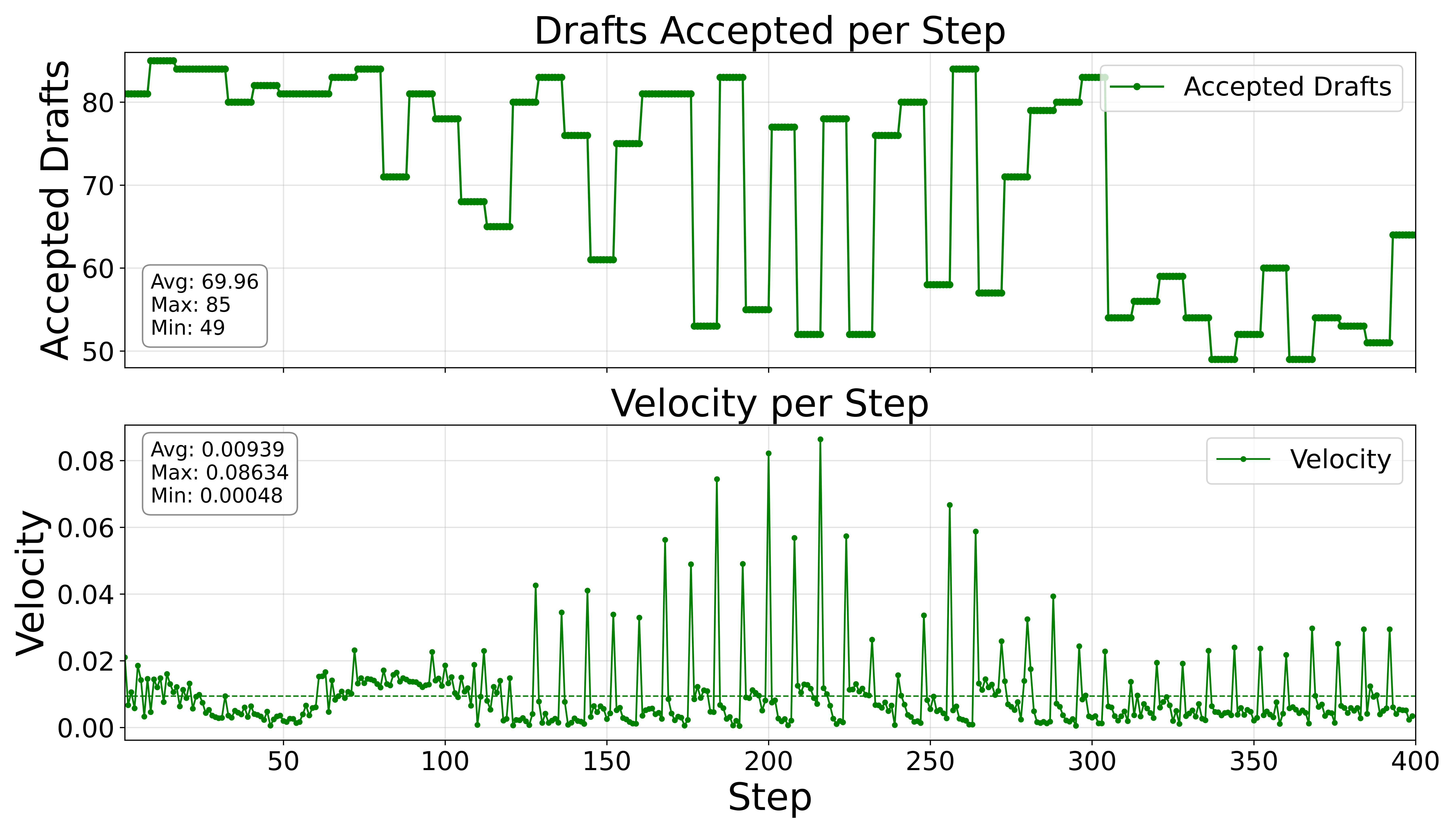}
    \caption{
        \textbf{Effect of End-Effector Velocity on the Number of Accepted Drafts.}
        Variation of Accepted Drafts and Velocity During the Can-ph Task: (Top) Accepted Drafts per Step; (Bottom) Velocity per Step.
        }
    \label{fig:400step}
\end{figure}

Assuming a Gaussian conditional distribution, the Metropolis–Hastings log acceptance ratio for the $i$-th draft proposal $\mathbf{x}=\hat{\boldsymbol{\mu}}_i+\sigma_i\boldsymbol{\xi}_i$, where $\boldsymbol{\xi}_i\sim\mathcal{N}(\mathbf{0},\mathbf{I})$, is:
\begin{equation}
    \log \alpha_i
    =
    -\tfrac{1}{2}
    \left\|
        \frac{\hat{\boldsymbol{\mu}}_i - \boldsymbol{\mu}_i}{\sigma_i}
    \right\|_2^2
    -
    \left\langle
        \frac{\hat{\boldsymbol{\mu}}_i - \boldsymbol{\mu}_i}{\sigma_i},
        \boldsymbol{\xi}_i
    \right\rangle.
\end{equation}
The corresponding acceptance probability is computed as:
\begin{equation}
    p_i = \min\!\left\{1,\, \exp(\log \alpha_i)\right\}.
\end{equation}
We accept the draft if $p_i \ge \lambda$, where $\lambda \in (0, 1]$ is a tunable acceptance threshold; otherwise, it is rejected. 
As shown in Fig.~\ref{fig:400step}, the number of accepted drafts is generally inversely related to the end-effector velocity—more drafts are accepted during slower motions, while fewer are accepted at higher speeds. 
This observation aligns with the typical behavior of embodied agents, which perform coarse, high-speed actions such as locomotion and fine, low-speed actions such as grasping \cite{li2025sp}. 
However, the non-strict correlation between velocity and acceptance rate indicates that task complexity also plays a critical role. 
Hence, using fixed speculative decoding parameters is suboptimal for embodied control. 
To address this, we treat $\lambda$ as an adaptive hyperparameter tuned by the scheduler based on task complexity and velocity, enabling efficient and stable speculative decoding across task phases.

Finally, we scan the candidates in chronological order, committing all consecutive drafts that are accepted until the first rejection occurs.
For the first rejected draft, we apply the reflection coupling algorithm to construct a corrected sample that preserves the original stochasticity while better aligning with the target distribution.
Unlike standard speculative decoding under fixed proposal and target distributions, \sysname adaptively adjusts the effective variance, acceptance threshold, and draft budget; therefore, it targets comparable task-level performance rather than exact distributional equivalence with the original DP.

\subsection{Temporal-Complexity-Aware Speculative Decoding Scheduler}
\noindent \textbf{Markov Modeling for Speculative Decoding.}
To enable dynamic decision-making of speculative decoding parameters at each time step and train the RL model, we first formulate the speculative decoding process as a Markov chain with a fixed number of steps.
Specifically, we integrate the DP, drafter, and embodied simulation environment into a unified interactive system.
(1) \textbf{The observation space} comprises three components: the object states returned by the embodied environment (either images or state vectors), the embodied actions generated by DP at each time step, and the current task progress.
(2) \textbf{The action space} includes the sigma scale, acceptance threshold, and draft step configurations for the three distinct denoising stages.

We adopt a RL–based scheduler implemented with Proximal Policy Optimization (PPO)~\cite{ppo,furuta2019pixelrl,xu2020deep}, as it provides stable policy updates and efficient adaptation to continuous task variations. 
In terms of network design, we separately encode the observation, action, and process streams to prevent interference caused by their differing dimensionalities.
The extracted features are then aggregated using an MLP.
When the observation is an image, a CNN is employed for feature extraction; otherwise, an MLP is used.

\noindent \textbf{Reward function.} 
The reward function aims to balance accuracy and efficiency, and is therefore divided into two components: the final reward and the intermediate reward.
The final reward serves as the primary objective of the scheduler, aiming to ensure task accuracy.
It is categorized into two types: discrete outcomes and continuous outcomes:
\begin{equation}
    r_{\text{final}_{\text{dis}}} =
    \begin{cases}
        +R_{\text{final}}, & \text{success}, \\
        -R_{\text{final}}, & \text{failure}.
    \end{cases}
\label{eq:discrete_reward}
\end{equation}
\begin{equation}
    r_{\text{final}_{\text{con}}} = 2 \times R_{\text{final}} \times r_{{max}} - R_{\text{final}},
\label{eq:continuous_reward}
\end{equation}
where $R_{\text{final}}$ is a hyperparameter.
This distinction enables us to differentiate between completion-based tasks and binary success–failure tasks.

To transform the sparse rewards into dense ones, we design the intermediate reward as an efficiency metric:
\begin{equation}
    r_{\text{process}}^{(i)} = \left( \frac{n_{\text{accept}}}{n_{\text{draft}}} + \frac{n_{\text{accept}}}{n_{\text{diffusion}}} \right) \times \lambda,
\end{equation}
where $n_{\text{accept}}$, $n_{\text{draft}}$ and $n_{\text{diffusion}}$ denote the number of accepted drafts, the total number of drafts, and the number of diffusion steps per denoising cycle, respectively. 
$\lambda$ is a scaling factor that constrains the accumulated process reward to one-fourth of the final reward, encouraging the scheduler to pursue efficiency while maintaining accuracy:
\begin{equation}
    \lambda = \frac{R_{final}/4}{N_{\text{expected}}}, \quad N_{\text{expected}} = \left\lceil \frac{T_{\text{max}}}{\Delta t} \right\rceil,
\end{equation}
where $T_{\text{max}}$ is the max task steps, $\Delta t$ denotes the parameter update frequency.

Furthermore, to reduce computational overhead, the scheduler is executed in parallel with the encoder.
As the encoder is significantly larger in scale, the scheduler incurs no additional time cost.

\section{Experiments}
\label{sec:experiments}

\subsection{Experimental Settings}

\textbf{Simulation protocol.}
We evaluate \sysname on Robomimic, Push-T, Multimodal Block Pushing, and Kitchen. These benchmarks cover a broad range of manipulation skills: the Robomimic suite contains Lift (raising a cube above a height threshold), Can (pick-and-place of a soda can into a target bin), Square (inserting a square nut onto a peg with precise alignment), Transport (a bimanual task that hands an object between two arms and places it in a target bin), and Tool Hang (high-precision frame assembly and tool hanging); Push-T pushes a T-shaped block to a fixed target pose; Multimodal Block Pushing pushes two blocks into two target squares; and Franka Kitchen requires completing a sequence of household subtasks, testing long-horizon multi-stage control. Following the official Diffusion Policy protocol, we use the official checkpoints and training seeds, and evaluate each checkpoint with 50 environment initializations. Success rate is used for most tasks, while target-area coverage is used for Push-T and Block Pushing. In the tables, ``$x/y$'' denotes results from two official checkpoints.

\textbf{Baselines and implementation.}
We compare against DP~\cite{dp}, Frozen Target Draft~\cite{Bortoli2025AcceleratedDM}, SpeCa~\cite{speca}, and BAC~\cite{Ji2025BlockwiseAC} using the same target backbone, checkpoints, denoising steps, hardware, and evaluation conditions. Simulation experiments run on an NVIDIA A100 GPU with 40~GB memory. The scheduler takes observations, recent actions, and task progress as input, and adjusts the sigma scale, acceptance threshold $\lambda$, and three-stage draft budget $K$. The static ablation uses $K\in\{10,25,40\}$.

\begin{figure}[t]
    \centering
    \newlength{\pltfigH}\setlength{\pltfigH}{0.345\columnwidth}%
    \begin{minipage}[b]{0.47\linewidth}
        \centering
        \includegraphics[height=\pltfigH]{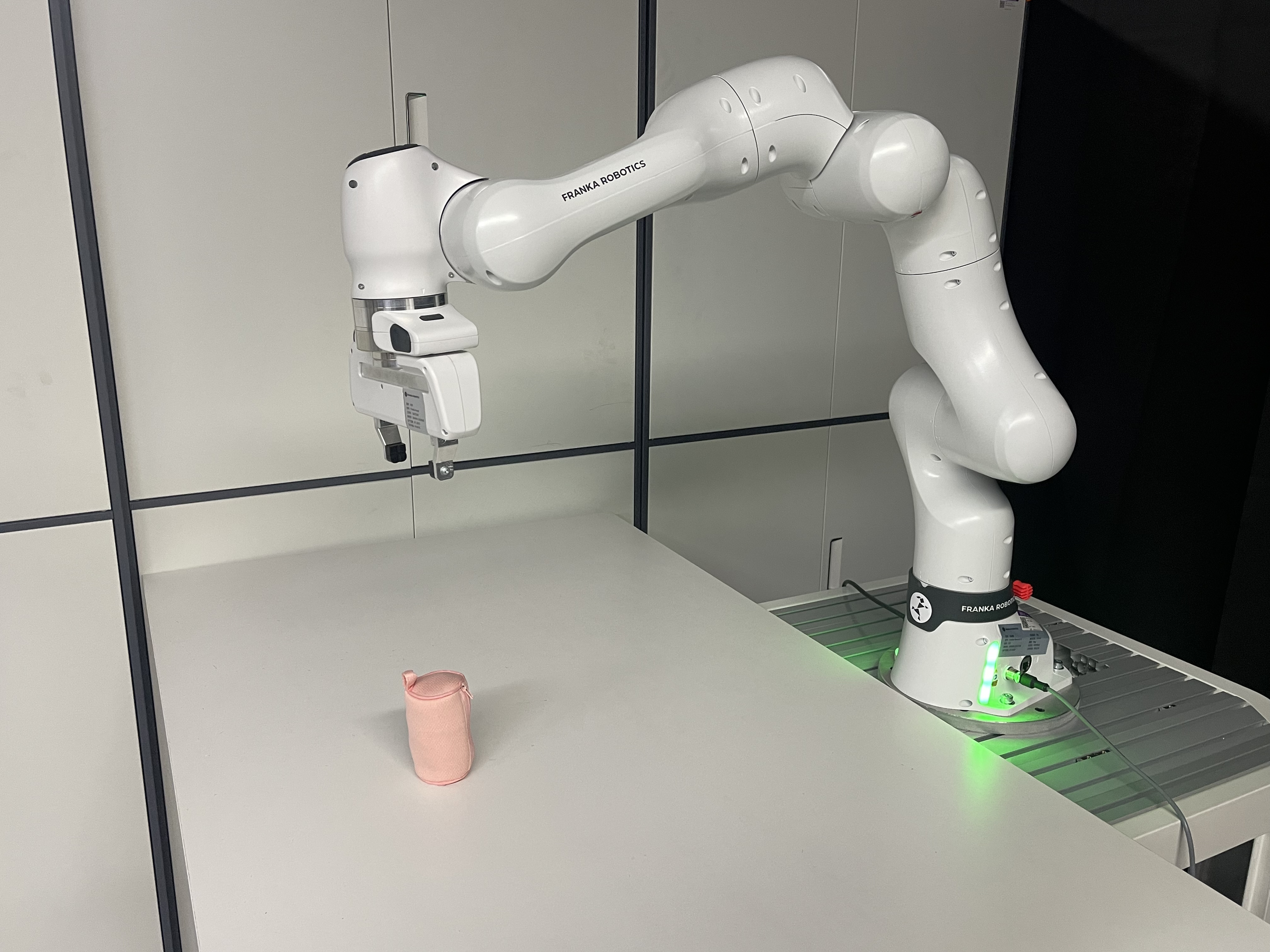}\\[-1pt]
        \small (a)
    \end{minipage}%
    \hspace{0.01\linewidth}%
    \begin{minipage}[b]{0.50\linewidth}
        \centering
        \includegraphics[height=\pltfigH]{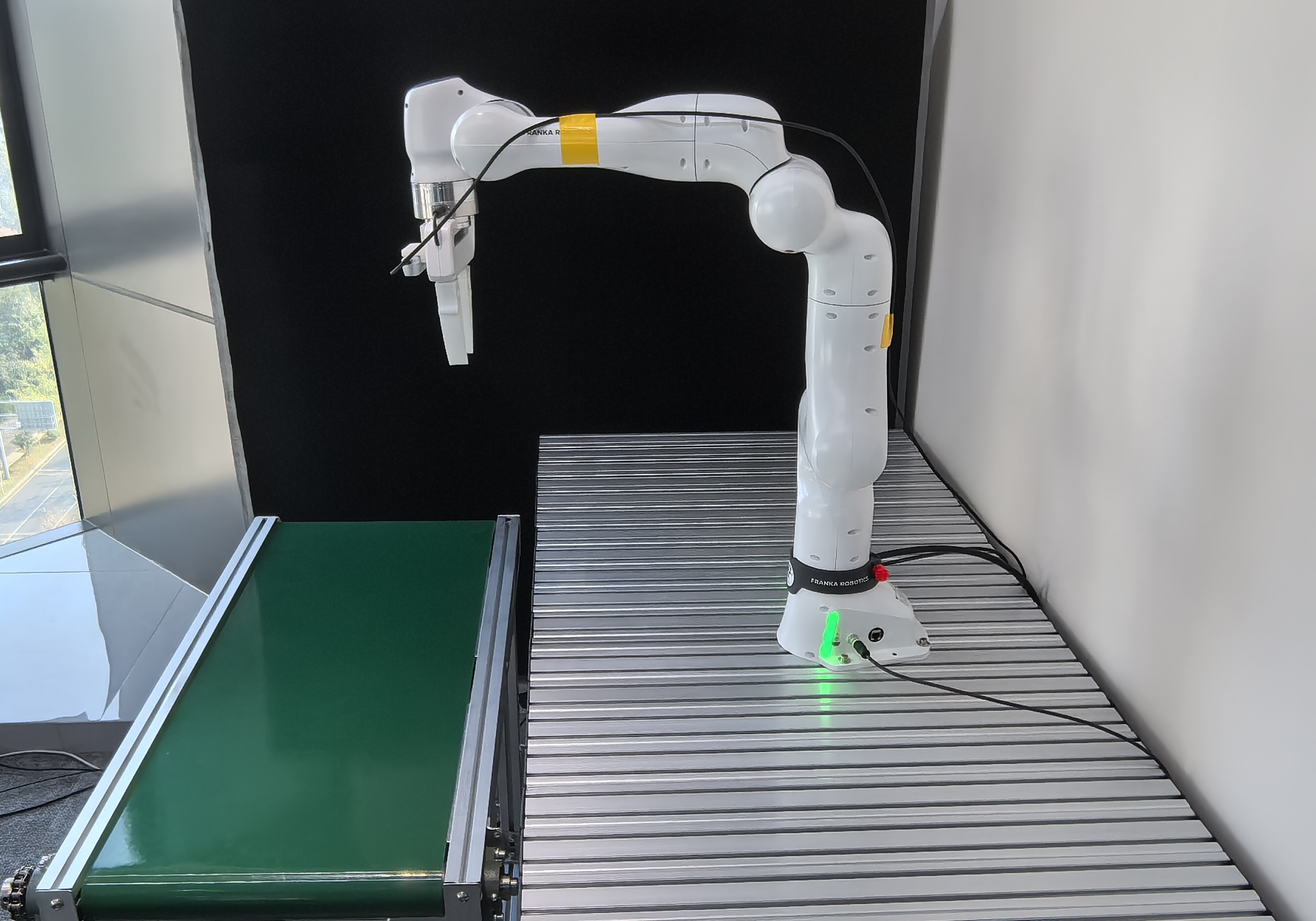}\\[-1pt]
        \small (b)
    \end{minipage}
    \caption{\textbf{Real-world experimental platforms.}
    The Pick-and-Release task is conducted on platform (a), while the
    Conveyor-Belt Pick-and-Place and Pen-to-Holder Insertion tasks are
    conducted on platform (b).}
    \label{fig:real-platforms}
\end{figure}

\textbf{Real-world setup.}
We use a Franka Research 3 arm and an Intel RealSense D435i camera in an eye-to-hand setting. Polymetis executes low-level control on a local NVIDIA RTX 4090 workstation, while the remote Diffusion Policy communicates through a socket interface. Each task is evaluated over 100 trials with randomized object placement within the reachable workspace. Failures mainly arise near workspace boundaries or under illumination changes. The experimental platforms are shown in Fig.~\ref{fig:real-platforms}.

\textbf{Real-world tasks.}
We design three tasks of increasing difficulty. \emph{Pick-and-Release} requires the arm to detect a pink cylinder, grasp it, lift it to a preset height, and release it; its difficulty is comparable to the official simulation and real-robot benchmarks. \emph{Conveyor-Belt Pick-and-Place} requires grasping a bun from a moving conveyor belt without stopping the belt and placing it onto a plate; this task is highly latency-sensitive and directly stresses inference speed. \emph{Pen-to-Holder Insertion} requires picking a pen from the table and inserting it vertically into a narrow-mouth holder that is not fixed and can be displaced by contact, a fine-manipulation task that challenges precision.

\begin{table}[t]
    \caption{\textbf{Results on real tasks.}}
    \centering
    \begin{threeparttable}
    {%
    \setlength{\tabcolsep}{5pt}
    \renewcommand{\arraystretch}{1.1}
    \begin{tabular*}{\linewidth}{@{\extracolsep{\fill}}lcccc@{}}
        \toprule
        \multirow{2}{*}{\textbf{Method}} &
        \multicolumn{3}{c}{\textbf{Success Rate (\%, $\uparrow$) / NFE ($\downarrow$)}} &
        \multirow{2}{*}{\textbf{AVG}} \\[-1pt]
        \cmidrule(lr){2-4}
        & Pick & Conveyor Belt & Pen Holder &  \\
        \midrule
        DP \cite{dp}
            & 82 / 100
            & 29 / 100
            & 31 / 100
            & 47 / 100 \\
        \textbf{\sysname}
            & 80 / \textbf{26}
            & \textcolor{red}{\textbf{52}} / \textbf{27}
            & 29 / \textbf{29}
            & \textcolor{red}{\textbf{54}} / \textcolor{blue}{\textbf{27}} \\
        \bottomrule
    \end{tabular*}
    }
    \begin{tablenotes}
        \footnotesize
        \item As task difficulty increases, the acceleration gain of \sysname{} decreases; moreover, the speed improvement is particularly beneficial for real-time conveyor-belt tasks, leading to a 23\% increase in success rate.
    \end{tablenotes}
    \end{threeparttable}
    \label{tab:real_tasks}
\end{table}

\begin{table*}[t]
    \centering
    \small
    \setlength{\tabcolsep}{6pt}
    \renewcommand{\arraystretch}{1.1}
    \caption{\textbf{Frequency and Latency Evaluation of \sysname.} Each entry reports frequency (Hz, $\uparrow$) / latency (s, $\downarrow$), with latency given as mean $\pm$ standard deviation over repeated trials.}
    \label{tab:frequency}
    \begin{threeparttable}
    \begin{tabular*}{\textwidth}{@{\extracolsep{\fill}}l r@{\extracolsep{0pt}$\,\pm\,$}l@{\extracolsep{\fill}} r@{\extracolsep{0pt}$\,\pm\,$}l@{\extracolsep{\fill}} r@{\extracolsep{0pt}$\,\pm\,$}l@{\extracolsep{\fill}} c@{}}
        \toprule
        \multirow{2}{*}{\textbf{Method}} &
        \multicolumn{6}{c}{\textbf{Frequency (Hz, $\uparrow$) / Latency (s, $\downarrow$)}} &
        \multirow{2}{*}{\textbf{AVG}} \\
        \cmidrule(lr){2-7}
        & \multicolumn{2}{c}{\textbf{Lift}} & \multicolumn{2}{c}{\textbf{Can}}
        & \multicolumn{2}{c}{\textbf{Square}} & \\
        \midrule
        DP \cite{dp}
        & $7.75 / 0.129$ & $0.004$
        & $7.52 / 0.133$ & $0.003$
        & $7.52 / 0.133$ & $0.004$
        & \multirow{6}{*}{%
            \begin{tabular}{c}
                \textbf{DP:} \\
                $8.15 / 0.124$ \\[0.4em]
                \textbf{\sysname:} \\
                $\mathbf{26.28 / 0.039}$
            \end{tabular}} \\
        \textbf{\sysname}
        & $\mathbf{24.39 / 0.041}$ & $\mathbf{0.002}$
        & $\mathbf{24.39 / 0.041}$ & $\mathbf{0.002}$
        & $\mathbf{23.81 / 0.042}$ & $\mathbf{0.003}$
        & \\
        \cmidrule{1-7}
        \multirow{2}{*}{\textbf{Method}} &
        \multicolumn{6}{c}{\textbf{Frequency (Hz, $\uparrow$) / Latency (s, $\downarrow$)}} &
        \\
        \cmidrule(lr){2-7}
        & \multicolumn{2}{c}{\textbf{Transport}} & \multicolumn{2}{c}{\textbf{Tool}}
        & \multicolumn{2}{c}{\textbf{Push-T}} & \\
        \cmidrule{1-7}
        DP \cite{dp}
        & $7.69 / 0.130$ & $0.005$
        & $8.13 / 0.123$ & $0.003$
        & $10.31 / 0.097$ & $0.003$
        & \\
        \textbf{\sysname}
        & $\mathbf{24.39 / 0.041}$ & $\mathbf{0.002}$
        & $\mathbf{25.00 / 0.040}$ & $\mathbf{0.002}$
        & $\mathbf{35.71 / 0.028}$ & $\mathbf{0.001}$
        & \\
        \bottomrule
    \end{tabular*}
    \begin{tablenotes}
        \footnotesize
        \item Compared with DP, \sysname{} increases the average control frequency by $3.2\times$, from 8.15~Hz to 26.28~Hz, while reducing the average latency from 0.124~s to 0.039~s. On the simplest Push-T task, the control frequency further reaches 35.71~Hz, well above the requirement for real-time control.
    \end{tablenotes}
    \end{threeparttable}
\end{table*}

\subsection{Main Results}
Here, we measure efficiency using Number of Function Evaluations (NFE) and inference speedup, where each drafter evaluation counts as 1/8 NFE and each target-model evaluation counts as 1 NFE.
NFE is a widely adopted efficiency metric in diffusion acceleration research~\cite{Lu2022DPMSolver,Wang2024OneStepDP}. Because DP inference is memory-bandwidth-bound, its latency is dominated by repeated sequential model calls rather than raw computation; \sysname reduces these sequential target calls through speculative generation and parallel verification. We therefore use NFE to characterize sequential sampling cost and do not interpret its reduction as a proportional reduction in FLOPs or MACs; the corresponding wall-clock gains are reported separately in the Latency Analysis.
Here, ``near-lossless'' refers only to empirically comparable success rates or target-area coverage relative to the original DP, rather than exact distributional equivalence.
Tables~\ref{tab:main_result_ph} and~\ref{tab:main_result_mh} present the acceleration performance of \sysname across different benchmarks. 
On the PH dataset, \sysname achieves a 4.17$\times$ inference speedup with 76\% reduction in NFE while achieving higher success rates of 85 / 80 compared to the baseline DP's 76 / 80. 
On the more challenging MH dataset, our method achieves a 3.84$\times$ speedup with 74\% NFE reduction while achieving success rates of 75 / 84 compared to the baseline's 76 / 81.
For complex multi-stage tasks, Table~\ref{tab:main_result_3} shows consistent acceleration benefits. 
On the Kitchen benchmark and Block Pushing tasks, \sysname achieves a 3.70\(\times\) speedup with 73\% NFE reduction while maintaining high performance, with an average success rate of 99 / 99 across subtasks.

Besides, we conduct real-world experiments under 3 settings: static pick-and-place, conveyor-belt pick-and-place, and placing a pen into a narrow-mouth holder. 
The quantitative results are reported in Table~\ref{tab:real_tasks}, while qualitative visualizations are provided in Figs.~\ref{fig:real_infer}, \ref{fig:pick_and_insert}, and \ref{fig:pick_and_release}.
On the simplest Pick-and-Release task, we achieve a 3.8$\times$ NFE speedup. 
For the real-time–critical conveyor-belt picking task, \sysname improves the success rate by 23\% while maintaining a 3.7$\times$ NFE speedup, demonstrating that improved real-time performance is crucial for dynamic tasks. 
Even on the most challenging fine-manipulation task, \sysname attains a 3.4$\times$ NFE speedup, indicating its ability to adapt to task difficulty while preserving acceleration gains.

\noindent \textbf{Action Fidelity.} 
To further characterize the near-lossless behavior beyond task success, we measure action-level fidelity between DP and \sysname. Under the same observation, official checkpoint, initial noise, and random seed, we generate paired action chunks and compute the normalized root mean squared error (NRMSE) and cosine similarity in the normalized action space. As shown in Table~\ref{tab:action-fidelity}, NRMSE stays below 0.05 and cosine similarity above 0.93 on all six tasks, with averages of 0.0318 and 0.9574, respectively. This confirms that \sysname closely reproduces the target policy's actions in practice, supporting empirical fidelity rather than exact distributional equivalence.

\begin{table}[t]
    \centering
    \small
    \setlength{\tabcolsep}{4pt}
    \renewcommand{\arraystretch}{1.1}
    \caption{\textbf{Empirical action-level fidelity between DP and \sysname,} reported as mean $\pm$ standard deviation.}
    \label{tab:action-fidelity}
    \begin{threeparttable}
    \begin{tabular*}{\linewidth}{@{\extracolsep{\fill}}lcc@{}}
        \toprule
        \textbf{Task} & \textbf{NRMSE $\downarrow$} & \textbf{Cosine Similarity $\uparrow$} \\
        \midrule
        Can       & $0.0324 \pm 0.0015$ & $0.9527 \pm 0.0082$ \\
        Lift      & $0.0384 \pm 0.0018$ & $0.9455 \pm 0.0091$ \\
        Square    & $0.0434 \pm 0.0020$ & $0.9387 \pm 0.0095$ \\
        Transport & $0.0454 \pm 0.0021$ & $0.9304 \pm 0.0103$ \\
        Tool      & $0.0101 \pm 0.0005$ & $0.9958 \pm 0.0021$ \\
        Push-T    & $0.0213 \pm 0.0010$ & $0.9814 \pm 0.0020$ \\
        \midrule
        \textbf{Average} & $\mathbf{0.0318}$ & $\mathbf{0.9574}$ \\
        \bottomrule
    \end{tabular*}
    \begin{tablenotes}
        \footnotesize
        \item Fidelity is slightly lower on contact-rich tasks such as Square and Transport, mirroring the small success-rate differences above; the consistently low NRMSE and high cosine similarity indicate that \sysname reproduces the target policy's actions rather than merely matching final task outcomes.
    \end{tablenotes}
    \end{threeparttable}
\end{table}

\subsection{Ablation Study}

\begin{table}[t]
    \caption{\textbf{Drafter-Only Results on the Mixed Human (MH) Demonstration Dataset.}}
    \centering
    \begin{threeparttable}
    {%
    \setlength{\tabcolsep}{5pt}
    \renewcommand{\arraystretch}{1.1}
    \begin{tabular*}{\linewidth}{@{\extracolsep{\fill}}lccccc@{}}
        \toprule
        \multirow{2}{*}{\textbf{Method}} &
        \multicolumn{4}{c}{\textbf{Success Rate (\%, $\uparrow$)}} &
        \multirow{2}{*}{\textbf{AVG}} \\[-1pt]
        \cmidrule(lr){2-5}
        & Lift & Can & Square & Transport &  \\
        \midrule
        DP \cite{dp}
            & 99 / 100
            & 92 / 97
            & 76 / 79
            & 35 / 46
            & 75.5 / 80.5 \\
        Drafter Only
            & 36 / 40
            & 26 / 20
            & \textcolor{blue}{0 / 0}
            & \textcolor{blue}{0 / 0}
            & \textcolor{blue}{15.5 / 15.0} \\
        \textbf{\sysname}
            & \textbf{99 / 99}
            & \textbf{92 / 97}
            & \textbf{74 / 78}
            & \textbf{34 / 62}
            & \textbf{74.75 / 84} \\
        \bottomrule
    \end{tabular*}
    }
    \begin{tablenotes}
        \footnotesize
        \item Using only the Drafter causes severe performance degradation, especially on harder tasks, whereas \sysname maintains high accuracy through speculative decoding and the PPO scheduler.
    \end{tablenotes}
    \end{threeparttable}
    \label{tab:drafter_only}
\end{table}

\noindent \textbf{Latency Analysis.} 
We first empirically justify the block-equivalent NFE definition. The target backbone contains eight Transformer blocks whereas the drafter contains one block of the same width; profiling both under identical inputs (Table~\ref{tab:block-latency}) yields average latencies of 9.37~ms and 1.17~ms, an $8.01\times$ ratio that supports counting one drafter call as $1/8$ target-equivalent NFE. This is an architecture-normalized measure of sampling cost and is reported separately from the wall-clock results below.

\begin{table}[t]
    \centering
    \small
    \setlength{\tabcolsep}{4pt}
    \caption{\textbf{Block-level latency (ms) of the target backbone and the drafter,} averaged over repeated runs.}
    \label{tab:block-latency}
    \resizebox{\linewidth}{!}{%
    \begin{tabular}{lccccccc}
        \toprule
        \textbf{Model} & \textbf{Lift} & \textbf{Can} &
        \textbf{Square} & \textbf{Transport} & \textbf{Tool} &
        \textbf{Push-T} & \textbf{Avg} \\
        \midrule
        DP &
        10.03 & 9.57 & 10.14 & 10.46 & 7.88 & 8.15 & 9.37 \\
        Drafter &
        1.22 & 1.19 & 1.19 & 1.25 & 1.15 & 1.02 & 1.17 \\
        \midrule
        \textbf{Ratio} &
        $8.22\times$ & $8.04\times$ & $8.52\times$ &
        $8.37\times$ & $6.85\times$ & $7.99\times$ &
        \textbf{$8.01\times$} \\
        \bottomrule
    \end{tabular}%
    }
\end{table}
We further evaluate the practical inference latency of \sysname, measuring the wall-clock time required for single action prediction across different robotic manipulation tasks. 
As reported in Table~\ref{tab:frequency}, the baseline DP requires 0.124s per action prediction, limiting its applicability in high-frequency control scenarios. 
Our \sysname framework reduces this latency to 0.039s, achieving a 3.2$\times$ speedup while maintaining identical or better task performance. 
Notably, \sysname achieves an average prediction frequency of 26.28 Hz compared to DP's 8.15 Hz, well exceeding the minimum requirement for real-time robotic control. 
The gain is also consistent across tasks of different difficulty: on simpler and more predictable tasks such as Push-T, the control frequency can be increased to 35.71 Hz, while on more challenging manipulation settings the system still maintains roughly 24 Hz operation. 
This indicates that \sysname does not merely provide a fixed acceleration factor, but instead adapts its speculative behavior to the temporal complexity of each task phase, allocating more aggressive parallel decoding when the denoising trajectory is stable and becoming more conservative near precision-critical moments. 
Overall, this latency profile enables \sysname to meet the timing requirements of real-time robotic control without compromising decision quality, making it suitable for practical deployment on edge devices with limited computational resources.

\noindent \textbf{Drafter Analysis.} 
We further evaluate a Drafter-only baseline to determine whether the gains of \sysname originate solely from replacing DP with a lightweight model. 
As shown in Table~\ref{tab:drafter_only}, using only the Drafter leads to a severe performance drop, and the degradation becomes more pronounced as task difficulty increases. 
Even on the relatively simple Lift task, the success rate falls to around 40\%, while on more challenging tasks such as Square and Transport it collapses to 0\%. 
These results indicate that the Drafter alone cannot provide reliable embodied control due to its limited representational capacity and weaker robustness at critical manipulation stages. 
This behavior is expected in embodied control, where task success often depends on a few key moments such as grasp closure, object alignment, or contact-rich transitions; once model capacity is reduced, these fine-grained decisions become much less reliable. 
As task difficulty increases, the weaker generalization ability of the compact model further amplifies this issue, causing errors that cannot be corrected once execution enters a critical stage. 
In contrast, \sysname preserves the accuracy of the full policy, showing that its effectiveness comes from the combination of the Drafter, speculative verification, and adaptive scheduling rather than from model shrinking alone.

\begin{figure}[t]
    \centering
    \includegraphics[width=\linewidth]{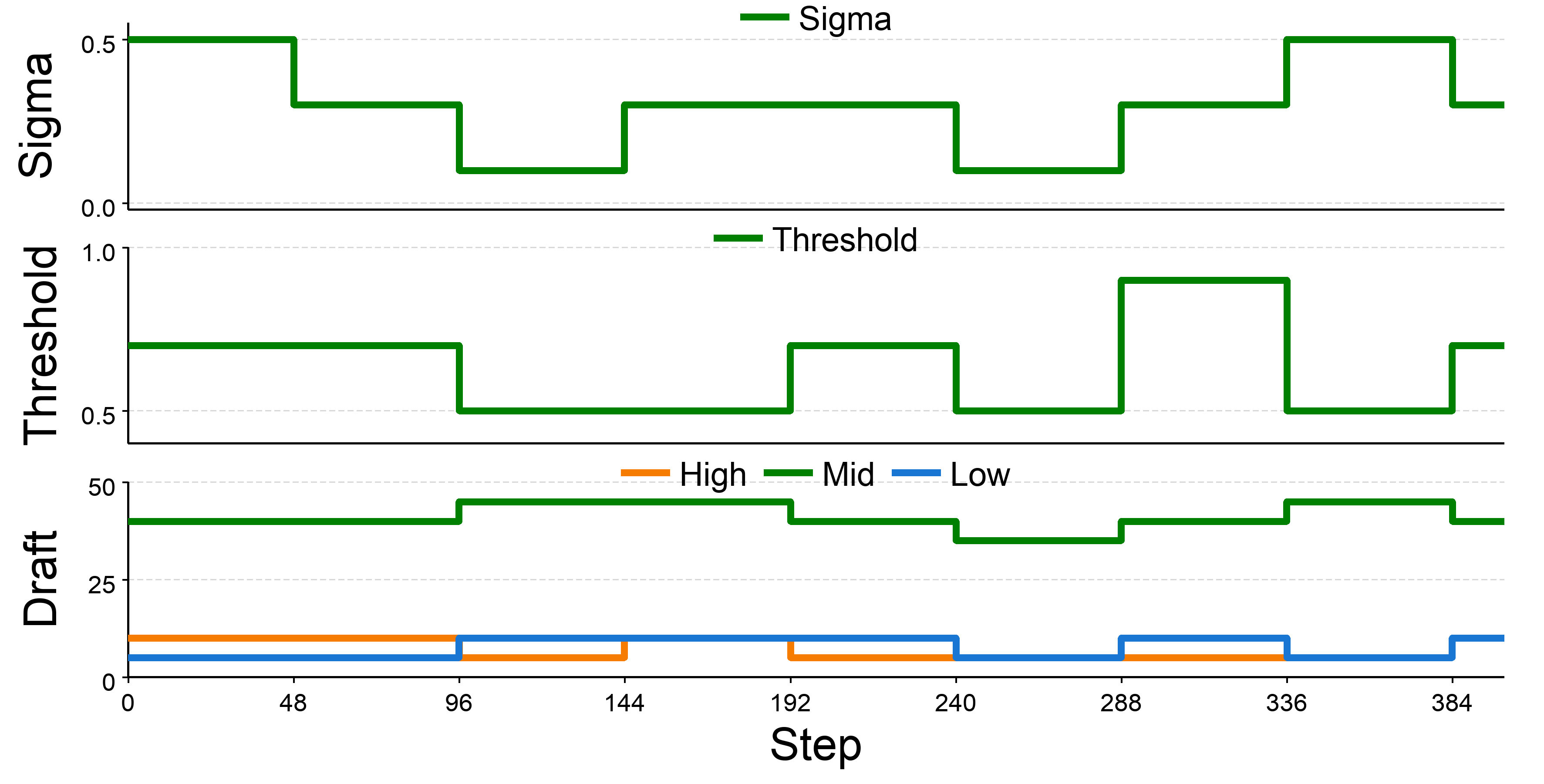}
    \caption{\textbf{Temporal scheduling dynamics on the Square task.} The scheduler adjusts the effective variance (Sigma), acceptance threshold, and the high-, intermediate-, and low-noise draft budgets over execution steps.}
    \label{fig:sched}
\end{figure}

\noindent \textbf{Effectiveness of Adaptive Scheduling.}
\begin{table}[t]
    \caption{\textbf{Performance Comparison With and Without the Scheduler on the Mixed Human (MH) Demonstration Dataset.}}
    \centering
    \begin{threeparttable}
    {%
    \setlength{\tabcolsep}{5pt}
    \renewcommand{\arraystretch}{1.1}
    \begin{tabular*}{\linewidth}{@{\extracolsep{\fill}}lcccccc@{}}
        \toprule
        \multirow{2}{*}{\textbf{Method}} &
        \multicolumn{4}{c}{\textbf{Success Rate (\%, $\uparrow$)}} &
        \multirow{2}{*}{\textbf{AVG}} &
        \multirow{2}{*}{\textbf{Speed $\times$}} \\[-1pt]
        \cmidrule(lr){2-5}
        & Lift & Can & Square & Transport &  &  \\
        \midrule
        $K=10$
            & 100
            & 94
            & 76
            & 66
            & 84
            & 2.45 \\
        $K=25$
            & 88
            & 88
            & 62
            & 60
            & 75
            & 3.47 \\
        $K=40$
            & 74
            & 90
            & 64
            & 60
            & 72
            & 3.92 \\
        \textbf{\sysname}
            & \textbf{100}
            & \textbf{96}
            & \textbf{84}
            & \textbf{66}
            & \textbf{87}
            & \textbf{3.80} \\
        \bottomrule
    \end{tabular*}
    }
    \begin{tablenotes}
        \footnotesize
        \item Fixed draft counts expose a clear speed--accuracy trade-off, whereas the adaptive scheduler achieves the best overall success rate while retaining high acceleration.
    \end{tablenotes}
    \end{threeparttable}
    \label{tab:ablation_mh_main}
\end{table}
To validate the effectiveness of our scheduler, we compare \sysname with static speculative decoding settings that keep the draft step count $K$ fixed throughout denoising. 
As shown in Table~\ref{tab:ablation_mh_main}, static configurations inevitably produce a trade-off between task success and acceleration: conservative decoding ($K=10$) yields stronger accuracy but limited speedup, while aggressive decoding ($K=40$) accelerates more but significantly hurts performance. 
In contrast, our adaptive scheduler achieves the best average success rate of 87\% while still reaching a 3.80$\times$ speedup. 
Compared with the best fixed setting, the adaptive policy improves the average success rate by 3 points over $K=10$ while preserving nearly the same acceleration regime as the more aggressive settings. 
This result suggests that a single static draft budget is fundamentally mismatched to embodied manipulation: stable phases can benefit from larger speculative batches, whereas contact-rich or alignment-sensitive phases require smaller budgets and stricter verification. 
Building on this observation, temporally varying embodied tasks cannot be effectively handled with a fixed speculative configuration. In contrast, our RL-based scheduler dynamically adjusts the decoding behavior to balance efficiency and reliability across different task phases.
Fig.~\ref{fig:sched} visualizes this behavior on the Square task, showing how the scheduler varies the effective variance (Sigma), the acceptance threshold, and the high-, intermediate-, and low-noise draft budgets over execution steps. It enlarges the intermediate-stage budget during stable phases and tightens the acceptance threshold while shrinking the high- and low-noise budgets during interaction-critical phases, confirming that the PPO scheduler adapts to task phase rather than using a fixed configuration.

\begin{table}[t]
    \caption{\textbf{Results Using a Speed Threshold on the Mixed Human (MH) Demonstration Dataset.}}
    \centering
    \begin{threeparttable}
    \scriptsize
    {%
    \setlength{\tabcolsep}{5pt}
    \renewcommand{\arraystretch}{1.1}
    \begin{tabular*}{\linewidth}{@{\extracolsep{\fill}}lcccccc@{}}
        \toprule
        \multirow{2}{*}{\textbf{Method}} &
        \multicolumn{4}{c}{\textbf{Success Rate / Draft Accept Rate (\%, $\uparrow$)}} &
        \multirow{2}{*}{\textbf{AVG}} &
        \multirow{2}{*}{\textbf{NFE ($\downarrow$)}} \\[-1pt]
        \cmidrule(lr){2-5}
        & Lift & Can & Square & Transport &  &  \\
        \midrule
        DP \cite{dp}
            & 99.5 / --
            & 94.5 / --
            & 77.5 / --
            & 40.5 / --
            & 78.0 / --
            & 100 \\
        VT \cite{li2025sp}
            & \textcolor{blue}{90 / 80}
            & \textcolor{blue}{86 / 80}
            & \textcolor{blue}{61 / 79}
            & \textcolor{blue}{35 / 81}
            & \textcolor{blue}{68 / 80}
            & 36 \\
        \textbf{\sysname}
            & \textbf{99 / 84.2}
            & \textbf{94.5 / 87.9}
            & \textbf{76 / 86.0}
            & \textbf{48 / 89.6}
            & \textbf{79 / 86.9}
            & \textbf{26} \\
        \bottomrule
    \end{tabular*}
    }
    \begin{tablenotes}
        \footnotesize
        \item VT means using a velocity threshold heuristic for scheduling \cite{li2025sp}.
        Replacing the PPO scheduler with a velocity-threshold heuristic reduces both average task success and acceptance quality, despite still lowering NFE.
    \end{tablenotes}
    \end{threeparttable}
    \label{tab:speed_threshold_mh_main}
\end{table}

\begin{figure}[t]
    \centering
    \includegraphics[width=\linewidth]{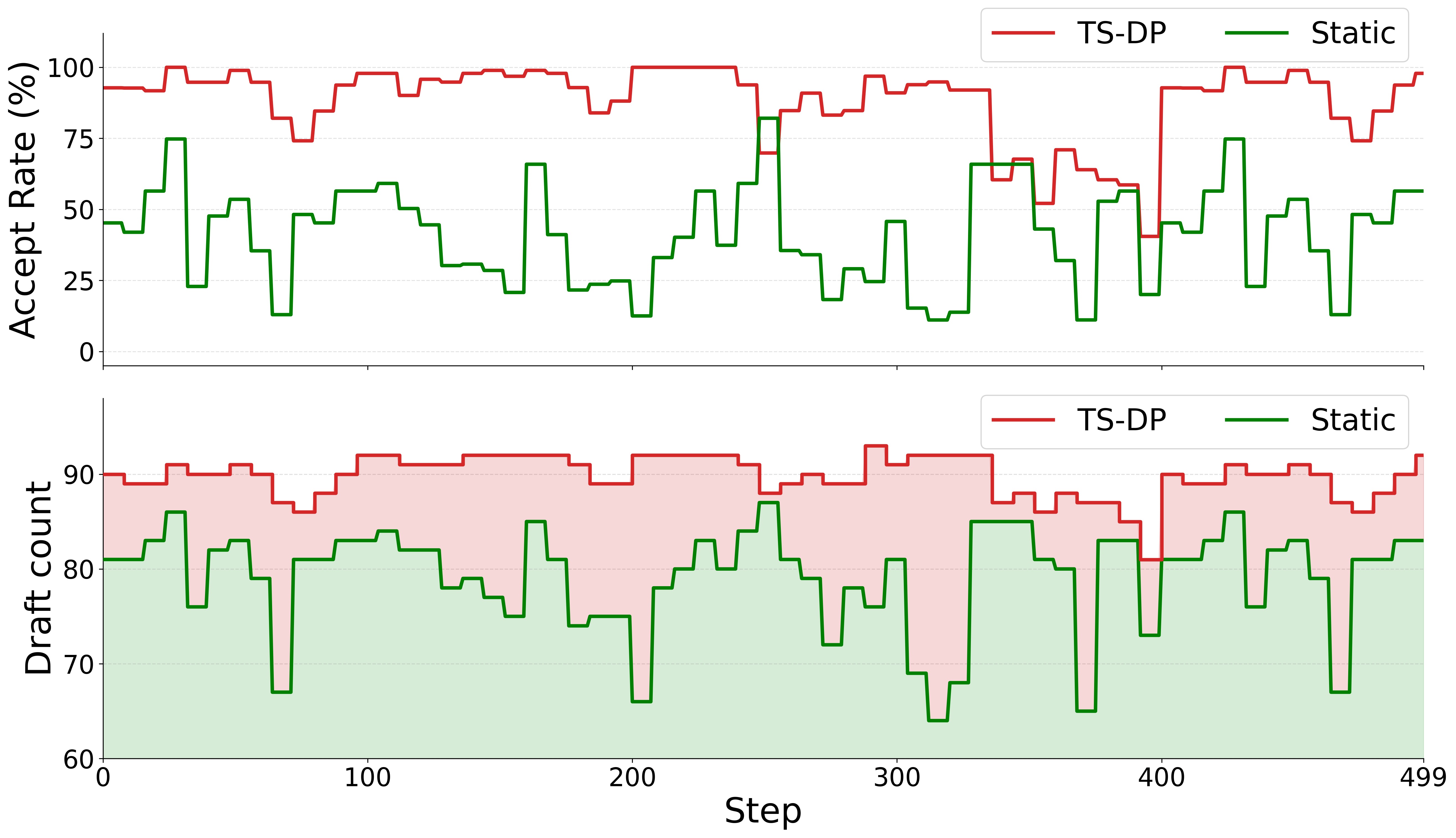}
    \caption{
        \textbf{Effect of \sysname on Acceptance Rate and Draft Count.}
        (Top) Acceptance Rate; (Bottom) Draft  Count. With the integration of \sysname, more drafts are accepted while maintaining task accuracy.
        }
    \label{fig:accept_and_draft}
\end{figure}
\begin{table}[t]
    \centering
    \small
    \setlength{\tabcolsep}{4pt}
    \caption{\textbf{Speculative decoding statistics across benchmarks.} We report NFE, speedup, draft count, and acceptance rate on the PH, MH, and Kitchen/BP benchmarks.}
    \label{tab:speculative-stats}
    \begin{tabular*}{\linewidth}{@{\extracolsep{\fill}}lcccc@{}}
        \toprule
        \textbf{Benchmark} & \textbf{NFE} & \textbf{Speed} &
        \textbf{Draft Count} & \textbf{Accept.\ Rate} \\
        \midrule
        PH & 24.0 & $4.17\times$ & 93.9 & 85.2\% \\
        MH & 26.3 & $3.83\times$ & 92.5 & 87.1\% \\
        Kitchen/BP & 27.2 & $3.77\times$ & 91.8 & 89.8\% \\
        \bottomrule
    \end{tabular*}
\end{table}
\noindent \textbf{Results Using Speed Threshold.}
We further replace the PPO scheduler with a velocity-threshold heuristic following SP-VLA to test whether a hand-crafted rule is sufficient for adaptive control. 
As reported in Table~\ref{tab:speed_threshold_mh_main}, this heuristic lowers the average success rate to 68\%, which is 11 points below \sysname, and also yields consistently weaker draft acceptance across tasks. 
This result indicates that using velocity alone captures only coarse task progress and cannot reliably identify critical moments that require higher precision. 
In practice, motion speed may already be high when the robot is about to enter a delicate alignment or contact stage, so a heuristic that keeps acceleration aggressive at that moment can easily induce unrecoverable errors. 
By contrast, the PPO scheduler in \sysname leverages richer multi-modal observations to infer task phase more accurately, enabling stronger end-to-end acceleration without sacrificing task completion.

\noindent \textbf{Visualizations.}
Figs.~\ref{fig:real_infer}, \ref{fig:pick_and_insert}, and \ref{fig:pick_and_release} visualize the three real-world tasks considered in our evaluation. 
For example, Fig.~\ref{fig:real_infer} illustrates the conveyor-belt pick-and-place task, where objects are grasped from a moving conveyor belt and placed onto a plate. 
In this case, DP operates significantly slower than \sysname and exhibits prolonged pauses after grasping, resulting in a total completion time of 63s, whereas \sysname completes the task in 20s. Moreover, faster inference and more responsive interaction with the environment enable \sysname to increase the success rate by 23\%.
As shown in Fig.~\ref{fig:accept_and_draft}, \sysname yields significantly higher draft acceptance rates and more efficient parallel computation compared to static methods. 
Table~\ref{tab:speculative-stats} further summarizes these statistics: as task complexity grows, the scheduler reduces the draft count and NFE rises from 24.0 to 27.2 while the acceptance rate stays above 85\%, indicating that \sysname adjusts its speculative intensity to task difficulty rather than using a fixed configuration.
The red curves (with scheduling) consistently maintain higher acceptance while generating more drafts, demonstrating optimal trade-off between acceleration and accuracy. 
More importantly, the visualization reveals a clear phase-dependent pattern: during stable execution segments, the scheduler increases the draft budget to maximize parallelism, while in interaction-critical periods it reduces the number of drafts and tightens the decision criterion to avoid error accumulation. 
This confirms that fixed-parameter speculative decoding is limited in embodied tasks with time-varying complexity, while our framework achieves superior end-to-end performance through intelligent hyperparameter adaptation.

\subsection{Training Cost and Reusability}
\sysname{} directly uses a frozen official DP checkpoint and trains a corresponding drafter and PPO scheduler for each checkpoint. The drafter is target-checkpoint specific, while the scheduler is target-policy-and-environment specific; changing the checkpoint, task distribution, environment, or robot embodiment therefore requires retraining or fine-tuning. Since the base DP itself does not support cross-task or cross-embodiment transfer, we make no additional generalization claim beyond the capability of the base model, and the real-robot experiments likewise use acceleration modules trained for the corresponding real-robot checkpoint.

All acceleration modules are trained offline on a single NVIDIA A40 GPU. Training the lightweight drafter uses 2{,}728~MiB of GPU memory and about 75~s per epoch for 150 epochs, totaling roughly 3.13~hours. Training the PPO scheduler over 4{,}032 environment timesteps takes about 2.75~hours and uses 12{,}992~MiB of GPU memory. Training both modules sequentially therefore requires approximately 5.88~hours with a peak memory of 12{,}992~MiB. This one-time offline cost excludes training of the official DP checkpoint and is not included in the reported online inference latency, so it is amortized over repeated deployments of the accelerated policy.

\section{Conclusion}
\label{sec:conclusion}
In this work, we present \sysname, a temporal-aware speculative decoding framework that accelerates Diffusion Policy by combining efficient draft generation with adaptive scheduling.
A distilled Transformer-based drafter replaces expensive base-model denoising calls across multiple steps, while an RL-based scheduler dynamically adjusts speculative parameters based on task difficulty to maintain accuracy and maximize efficiency.
Extensive experiments across 4 manipulation benchmarks demonstrate up to a \textbf{4.17$\times$} NFE reduction (\textbf{76\%} fewer NFE), a \textbf{3.2$\times$} wall-clock speedup, and real-time operation at \textbf{26 Hz} while maintaining comparable task performance in simulation. 
Real-world experiments further show up to a \textbf{3.8$\times$} NFE reduction and improved success rates on real-time tasks (e.g., a \textbf{23\%} gain on the conveyor-belt task).

\section*{Acknowledgments}
This work was supported in part by the National Natural Science Foundation of China (Grant No. 92467204, 62472249, 62571298, and 62402264), and in part by the Shenzhen Science and Technology Program (Grant No. KJZD20240903102300001).

\balance
\bibliographystyle{IEEEtran}
\bibliography{IEEEabrv,reference}

\end{document}